\documentclass{article}

\usepackage{microtype}
\usepackage{graphicx}
\usepackage{subfigure}
\usepackage{booktabs} % for professional tables

\usepackage{hyperref}

\usepackage[accepted]{icml2024}

\usepackage{amsmath}
\usepackage{amssymb}
\usepackage{mathtools}
\usepackage{amsthm}

\usepackage[capitalize,noabbrev]{cleveref}

\theoremstyle{plain}
\newtheorem{theorem}{Theorem}[section]

\theoremstyle{definition}
\newtheorem{definition}[theorem]{Definition}

\theoremstyle{remark}

\usepackage[textsize=tiny]{todonotes}

\icmltitlerunning{CDM for EMOBO}

\begin{document}

\twocolumn[
\icmltitle{Expensive 
Multi-Objective Bayesian Optimization
Based on 
% Conditional and Unconditional  
Diffusion Models}



\icmlsetsymbol{equal}{*}

\begin{icmlauthorlist}
\icmlauthor{Bingdong Li}{ECNU}
\icmlauthor{Zixiang Di}{ECNU}
\icmlauthor{Yongfan Lu}{ECNU}
\icmlauthor{Hong Qian}{ECNU}
\icmlauthor{Feng Wang}{WHU}
\icmlauthor{Peng Yang}{SUSTech}
\icmlauthor{Ke Tang}{SUSTech}
\icmlauthor{Aimin Zhou}{ECNU}
\end{icmlauthorlist}

\icmlaffiliation{ECNU}{East China Normal University}
\icmlaffiliation{WHU}{Wuhan University}
\icmlaffiliation{SUSTech}{Southern University of Science and Technology}

\icmlcorrespondingauthor{Feng Wang}{fengwang@whu.edu.cn}
\icmlcorrespondingauthor{Aimin Zhou}{amzhou@cs.ecnu.edu.cn}

\icmlkeywords{Machine Learning, ICML}

\vskip 0.3in
]



\printAffiliationsAndNotice{}  % leave blank if no need to mention equal contribution

\begin{abstract}
Multi-objective Bayesian optimization (MOBO) 
% \cite{laumanns2002bayesian}
has shown promising performance on various 
expensive multi-objective optimization problems (EMOPs). 
However,
effectively modeling complex distributions of
the Pareto optimal solutions is difficult
with limited function evaluations. 
Existing Pareto set learning  algorithms may exhibit considerable instability in such expensive scenarios,
leading to significant deviations between the obtained solution set and the Pareto set (PS). 
% In other words, the quality of the resulting solution set is highly influenced by the performance of the PSL model,
% which is largely limited on EMOBOPs. 
In this paper, we propose a novel
Composite Diffusion Model based Pareto Set Learning 
algorithm, namely CDM-PSL,
for expensive  MOBO.
CDM-PSL  includes both 
    unconditional and conditional diffusion model
    for generating high-quality samples.
Besides,
we introduce an information entropy based weighting  method to balance  different objectives of EMOPs.
This method is integrated with the guiding strategy, ensuring that   all the objectives are appropriately balanced and given due consideration  during the optimization process;
    % \item We design a switching strategy based on the predicted Hypervolume (HV) values to prevent the DM from falling into local optima. This algorithm switches the operators that populate the important feature dimensions.
    % \item \textcolor{red}{4) To be confirmed whether samples sort algorithm is enough to be a component? There is a self-written algorithm to sort all the samples, and then the DM selects the top third of the good samples for generation.}
Extensive experimental results on both synthetic benchmarks and real-world problems demonstrates that our proposed algorithm attains superior performance compared with various state-of-the-art MOBO algorithms.
% on EMOBOPs under a limited evaluation budget. 

% This document provides a basic paper template and submission guidelines.
% Abstracts must be a single paragraph, ideally between 4--6 sentences long.
% Gross violations will trigger corrections at the camera-ready phase.
% \textcolor{red}{ddl:
% As noted above, this year, ICML will use a single paper submission deadline with a single review cycle, as follows.
% \\
% Submissions open Jan 9th, 2024.
% \\
% Full paper submission deadline February 1st, 2024  AoE (Feb 02 2024 12 Noon UTC-0).
% \\
% Abstracts and papers can be submitted through OpenReview: https://openreview.net/group?id=ICML.cc/2024/Conference 
% \\
% Topics of interest include (but are not limited to):}
\end{abstract}

\section{Introduction}
 
Expensive multi-objective optimization problems  are commonly seen various fields, such as  neural architecture search \cite{zoph2018learning, lu2019nsga}, antenna structure design \cite{ding2019compact},
% exercise recommendation \cite{huang2019exploring}, 
and clinical drug trials \cite{yu2019simulation}. 
Handling EMOPs involves optimizing multiple (often conflicting) objectives simultaneously with the limit number of function evaluations due to time and financial constraints.

To meet these challenges, 
% the application of 
multi-objective Bayesian optimization (MOBO) \cite{laumanns2002bayesian}, an extension of single-objective Bayesian Optimization (BO) \cite{movckus1975bayesian} for expensive multi-objective optimization problems, has emerged as a promising paradigm. BO itself is recognized as an exceedingly effective strategy for global optimization, particularly noted for its success in addressing black-box optimization issues \cite{jones1998efficient, snoek2012practical}. 
The core principle of BO involves creating probabilistic surrogate models that closely represent the black-box functions. 
These models are utilized in conjunction with acquisition functions to seek out globally optimal solutions. 
MOBO represents a fusion of Bayesian optimization with multi-objective optimization. 
A widely adopted MOBO approach is the random scalarization technique, which effectively translates a multi-objective optimization problem  into several single-objective optimization problems.
Another noteworthy strategy in MOBO involves the use of sophisticated acquisition functions, such as the expected hypervolume improvement (EHVI) \cite{couckuyt2014fast} and predictive entropy search (PES) \cite{hoffman2015output}. 
Among them, Pareto set learning (PSL) based methods
(e.g.
\cite{lin2022pareto})
which aims to modeling the Pareto set via 
machine learning techniques, 
have shown promising performance.
% The goal of these algorithms is to find a set of approximate Pareto solutions with limited objective function evaluation budget.

However, effectively capturing and modeling complex distributions of limited samples is difficult when faced with expensive multi-objective Bayesian optimization problems (EMOBOPs). 
Existing PSL algorithms may exhibit considerable instability in expensive scenarios. 
This instability can lead to significant deviations between the obtained solution set and the true Pareto set (PS). 
In other words, the quality of the resulting solution set is highly influenced by the performance of the PSL model,
which is largely limited on EMOBOPs. 

Diffusion model (DM), inspired by the natural diffusion of gases, is a kind of popular deep generative models. 
The characteristics of DMs, including their distribution coverage, stationary training objective, and effortless scalability \cite{DMbeatGAN}, have empowered them to not only outperform Generative Adversarial Networks (GANs) in image synthesis tasks \cite{DMbeatGAN} but also achieve great success in diverse fields such as computer vision \cite{DM-CV1, DM-CV2}, natural language processing \cite{DM-NLP}, and waveform signal processing \cite{DM-Wave}. 
% Monte Carlo Tree Search (MCTS), a tree search algorithm based on random sampling, has marked milestone achievements beyond the realm of games such as Go \cite{GO1, GO2}. Its widespread application in robotics planning and optimization problems \cite{munos2014bandits, weinstein2012bandit, mansley2011sample} demonstrates its exceptional capability in the domain of spatial partitioning. 
These features of DM show promise for its application in Pareto set learning in EMOBOPs.

% MCTS-DM version:
% In this paper, we propose a novel expensive multi-objective Bayesian optimization based on Monte Carlo Tree Search and Diffusion Model, named MCTS-DM. We introduce the Diffusion Model into Pareto Set Learning, where DM operates by simulating the transition of data from an ordered state to disordered noise. This process progressively learns and reveals the inherent distribution within the data. This methodology is particularly effective in scenarios with limited sample sizes, as it can efficiently extract substantial information from each individual sample, thereby amplifying the learning impact \cite{yang2023diffusion}. This enables the effective modeling of complex distributions in high-quality samples. Furthermore, to acquire higher-quality samples, we integrate Monte Carlo Tree Search into our framework. MCTS is utilized to identify important features of the problem, thereby partitioning a low-dimensional subspace. This approach not only mitigates the curse of dimensionality but also enables the DM to more effectively learn the distribution of the Pareto optimal set within these significant feature dimensions.

% CDM version
In this paper, we propose a novel
Composite Diffusion Model based Pareto Set Learning
algorithm, namely CDM-PSL
% ()
for expensive 
MOBO.
% multi-objective Bayesian optimization algorithm 
 % Diffusion Model, named . 
 We introduce the Diffusion Model into Pareto Set Learning, where DM operates by simulating the transition of data from an ordered state to disordered noise. This process progressively learns and reveals the inherent distribution within the data. This methodology is particularly effective in scenarios with limited sample sizes, as it can efficiently extract substantial information from each individual sample, thereby amplifying the learning impact \cite{yang2023diffusion}. 
 This enables the effective modeling of complex distributions of high-quality samples. 
% Furthermore, to acquire higher-quality samples, we integrate Monte Carlo Tree Search into our framework. MCTS is utilized to identify important features of the problem, thereby partitioning a low-dimensional subspace. This approach not only mitigates the curse of dimensionality but also enables the DM to more effectively learn the distribution of the Pareto optimal set within these significant feature dimensions.
Building upon this foundation, to generate samples of superior quality, we designed a guided sampling process.  This approach ultimately led to the realization of 
conditional sample generation.
% Conditional Diffusion Model (CDM).

% The main contributions of this paper include: 1) We introduced diffusion model based Pareto set learning for offspring generation on EMOBOPs; 2) In order for DM to better predict the distribution of Pareto optimal sets, MCTS is utilized to select the important features of the problem. This also mitigates the curse of dimensionality; and 3) In order to avoid the DM from falling into a local optimum, a switching algorithm based on the predicted Hypervolume (HV) values is designed to switch the operators that populate the important feature dimensions. \textcolor{red}{and maybe 4) samples sort algorithm ?}

The major contributions of this paper are summarized as follows:
% \begin{itemize}
  1) We introduce a 
    composite diffusion model based Pareto set learning method for offspring generation 
    for expensive MOBO,
    which includes both 
    unconditional and conditional 
    sample generation.
    % \item We utilize Monte Carlo Tree Search to select important features of the problem. This enhances the ability of DM to better predict the distribution of Pareto optimal sets and also mitigates the curse of dimensionality.
2) We devise a guided sampling process to improve the quality of solutions generated by the diffusion model, resulting in a conditional diffusion model;
3) We introduce an information entropy based weighting  method to balance the importance of different objectives of multi-objective problems. This method is integrated with the guiding strategy, ensuring that 
    all the objectives are appropriately balanced and given due consideration 
    during the optimization process;
    % \item We design a switching strategy based on the predicted Hypervolume (HV) values to prevent the DM from falling into local optima. This algorithm switches the operators that populate the important feature dimensions.
    % \item \textcolor{red}{4) To be confirmed whether samples sort algorithm is enough to be a component? There is a self-written algorithm to sort all the samples, and then the DM selects the top third of the good samples for generation.}
4) We have conducted extensive experiments on both synthetic benchmarks and real-world problems, clearly demonstrating that CDM-PSL obtains superior performance compared with various state-of-the-art MOBO algorithms. 
    % Additionally, the effectiveness and efficiency of all the components have been conclusively validated.
% \end{itemize}

\section{Preliminaries}
\label{sec2related}

%%%%%%%%%%%%%%%%%%%%%%%%%%%%%%%%%%%%%%%%%%%%%%%%%%%%%%%%%%%%%%%%%%%%%%%%

% \textcolor{red}{current High dimensional MOBO methods?}\\
% \textcolor{red}{current Expensive MOBO methods?}\\
% \textcolor{red}{MCTS?}\\
% \textcolor{red}{CDM?}\\

%%%%%%%%%%%%%%%%%%%%%%%%%%%%%%%%%%%%%%%%%%%%%%%%%%%%%%%%%%%%%%%%%%%%%%%%

\subsection{Expensive Multi-objective Optimization}
A multi-objective optimization problem (MOP) can be universally expressed in mathematical terms as follows:
    \begin{eqnarray} \label{mopdef}
        &&\operatorname{minimize} \textbf{ } \boldsymbol{ f}(\boldsymbol{x})=(f_{1}(\boldsymbol{x}), f_{2}(\boldsymbol{x}),\ldots, f_M(\boldsymbol{x}))^{T}\nonumber\\
        &&subject\textrm{ }to\textrm{ }  \boldsymbol{x} \in \Omega    
    \end{eqnarray} 
where $\boldsymbol{x}$ $=$ $(x_1,x_2,\ldots,x_d)$ represents the decision vector,
% composed of $d$ variables. The function 
$\boldsymbol{ f} (\cdot) $: $\Omega\rightarrow\Lambda$ denotes a black-box objective function, encompassing $M$ ($M \geq 2$)   objectives,
% Here, 
$\Omega$ symbolizes the non-empty \textit{decision space}, 
and $\Lambda$ is the \textit{objective space}. An MOP is considered expensive when the evaluation of $\boldsymbol{f(x)}$ involves either time-intensive computations or high-cost experimental procedures. In such contexts, the primary aim in optimizing an MOP is to approximate the Pareto Front (PF) effectively within a limited evaluation budget.
\begin{definition}[Pareto dominance]
Considering two solutions $\boldsymbol{x}$ and $\boldsymbol{y}$ $\in$ $\Omega$,
% the term \textit{dominate} is used in the following context: 
$\boldsymbol{x}$ is said to \textit{dominate} $\boldsymbol{y}$ (expressed as $\boldsymbol{x} \prec \boldsymbol{y}$) if and only if the following conditions are met:
1) $\forall i \in \{1,2,...,M\}$, 
% $i$ in the set ${1, 2, \ldots, M}$, it holds that 
$f_{i}(\boldsymbol{x}) \leq f_{i}(\boldsymbol{y})$;
2) $\exists j \in \{1,2,...,M\}$,
% ?in the set ${1, 2, \ldots, M}$ 
 $f_{j}(\boldsymbol{x}) < f_{j}(\boldsymbol{y})$. 
 This definition encapsulates the essential criterion for determining 
 the quality of solutions in terms of meeting multiple objectives simultaneously. \cite{yu1974cone}
\end{definition} 
\begin{definition}[Pareto optimal]
A solution $\boldsymbol{x}^{\ast} \in \Omega$ is  Pareto optimal if there exists no other solution $\boldsymbol{x} \in \Omega$ that can dominate it. This implies that within the feasible region $\Omega$, $\boldsymbol{x}^{\ast}$ is considered to be Pareto optimal if no alternative solution offers better outcomes across all objectives without being worse in at least one of them.
\end{definition} 
\begin{definition}[Pareto Set and Pareto Front]
The \textit{Pareto Set} (PS) refers to the collection of all the Pareto optimal solutions:
$PS$$=$$\{\boldsymbol{x}  \in \Omega | \forall \boldsymbol{y}  \in \Omega,\boldsymbol{y} \not\prec  \boldsymbol{x}   \}$.
% In this context, the PS comprises every solution within the feasible region $\Omega$ that is not dominated by any other solution in the same space. Furthermore, 
The corresponding set of objective vectors of the PS is the \textit{Pareto Front} (PF).
% /
% , representing the frontier of these optimal trade-offs in the objective space.
\end{definition} 

\subsection{Bayesian Optimization}
Bayesian Optimization (BO) is a powerful method for the efficient global optimization of expensive black-box functions \cite{jones1998efficient, shahriari2015taking}. By leveraging probabilistic surrogate models to approximate the black-box functions, in conjunction with acquisition functions, it seeks to locate global optimal solutions with as few evaluations of the actual objective function as possible. BO has been widely used in a variety of fields, including hyperparameter tuning \cite{bergstra2011algorithms}, A/B testing \cite{letham2019constrained}, combinatorial optimization \cite{Zhang_2015_CVPR}, among others.

\subsection{Diffusion Models}
Diffusion models are a specialized form of probabilistic generative models that operate by learning to reverse a forward process that gradually increases noise in the training data \cite{sohl2015deep, ho2020denoising}.
They have demonstrated remarkable performance on a wide variety of tasks, such as image generation \cite{austin2021structured, bao2022analytic}, voice synthesis \cite{liu2022diffsinger}, video generation \cite{ho2022imagen} and inpainting \cite{lugmayr2022repaint}. 
Training a diffusion model involves two processes: the forward diffusion process and the backward denoising process.

\subsubsection{Forward Process}
In the forward phase, Gaussian noise is added to the input data step by step until a pure Gaussian noise is produced, which is a Markovian process. Given an initial data distribution $\mathbf{x}_0\sim q(\mathbf{x})$, the noised $x_1,x_2\ldots,x_T$ can be obtained from the following equation:
$$q(x_t|x_{t-1})=\mathcal{N}\Big(x_t;\sqrt{1-\beta_t}\cdot x_{t-1},\beta_t\cdot\mathbf{I}\Big),\forall t\in\{1,\ldots,T\},$$
where $T$ is the number of diffusion steps and the step sizes are controlled by a variance schedule $\{\beta_t\in(0,1)\}_{t=1}^t$. Moreover, the properties of this recursive formula make it possible to obtain $q(x_t)$ directly from $x_0$ by the following equation:
$$q(x_t|x_0)=\mathcal{N}\bigg(x_t;\sqrt{\hat{\beta}_t}\cdot x_0,(1-\hat{\beta}_t)\cdot\mathbf{I}\bigg), \forall t\in\{1,\ldots,T\},$$
where $\hat{\beta_t}=\prod_{i=1}^t\alpha_i$ and $\alpha_t=1-\beta_t$. Thus, $x_t$ can be sampled from $q(x_t|x_0)$ as follows:
$$x_t=\sqrt{\hat{\beta}_t}\cdot x_0+\sqrt{(1-\hat{\beta}_t)}\cdot z_t,$$
where $z_t\sim\mathcal{N}(0,\mathbf{I})$.

\subsubsection{Reverse Process}
The reverse process recreates the true sample from a Gaussian noise input $x_T\sim\mathcal{N}(0,\mathbf{I})$ by the following equation:
$$q(x_{t-1}|x_t)~=~\mathcal{N}(x_{t-1};\mu(x_t,t),\Sigma(x_t,t))$$
However, $q(x_{t-1}|x_t)$ cannot easily be evaluated due to the reverse process lacking a complete dataset and therefore we need to train a neural network $p_{\theta}(x_{t-1}|x_t)=\mathcal{N}(x_{t-1};\mu_\theta(x_t,t),\Sigma_\theta(x_t,t))$ to approximate these conditional probabilities. Specifically, The model takes in the noisy data $x_t$ and the corresponding embedding at time step $t$ as input, and is trained to predict the mean $\mu_\theta(x_t,t)$ and the covariance $\Sigma_\theta(x_t,t))$. Based on this, Ho \cite{ho2020denoising} proposed to fix the covariance $\Sigma_\theta(x_t,t))$ to a constant value and reformulating the mean $\mu_\theta(x_t,t)$ as a function dependent on noise, as follows:
$$\mu_\theta=\frac{1}{\sqrt{\alpha_t}}\cdot\left(x_t-\frac{1-\alpha_t}{\sqrt{1-\hat{\beta}_t}}\cdot z_\theta(x_t,t)\right).$$
This enables the model to 
predict the noise of the data rather than directly predicting the mean and the covariance.

\subsection{Conditional Generative Models}
In the field of generative models, Conditional Generative Models have garnered significant interest \cite{odena2017conditional, gong2019twin}. A prominent approach in Generative Adversarial Networks (GANs) involves incorporating a classification component within the Discriminator network. The strategy aims to facilitate the learning of classifier conditions, a method explored in various studies \cite{zhao2020feature, kang2021rebooting, dung2022gdegan}. The training of these classifiers proceeds occurs concurrently with the GAN training process.

Similarly, Deep Generative Models (DGMs) have been adapted to include conditional capabilities. This is achieved by integrating a classification head into diffusion models, as demonstrated in recent work \cite{song2020score}. There are instances where DGMs are conditioned on images, enabling style transfer across different instances, a concept explored by Preechakul \cite{preechakul2022diffusion}. The research by Dhariwal and Nichol \cite{dhariwal2021diffusion} presents a novel method allowing for the control of samplers/generators in DGMs without necessitating retraining. This idea is further expanded by Liu et al. \cite{liu2023more}, who generalize this concept to accommodate various modalities. In the realm of guidance without classifiers, Ho and Salimans \cite{ho2022classifier} have shown that it's possible to achieve guidance properties in generative models without relying on a classifier.

\subsection{Diffusion Model Based Optimization Algorithms}
Krishnamoorthy proposed a black-box optimization algorithm named DDOM \cite{krishnamoorthy2023diffusion}, based on the diffusion model. This algorithm converts a single-objective optimization problem into a continuous diffusion process, leveraging the inverse process of the diffusion model to efficiently address complex problems. Subsequently, Yan and Jin's EmoDM \cite{yan2024emodm} extended the application of the diffusion model to multi-objective optimization. By learning the noise distribution in the previous evolutionary search task, a set of non-dominated solutions can be generated for the new multi-objective optimization problem without further evolutionary search. Fang's DMO \cite{fang2024diffusion} further demonstrates the diffusion model's efficacy by applying it to create a gasoline hybrid scheduling scheme, highlighting its capability in solving practical multi-objective optimization challenges. Building upon these contributions, this paper's CDM-PSL advances the field by optimizing the balance between solution convergence and diversity through the integration of conditional and unconditional diffusion models. Moreover, CDM-PSL incorporates gradient information, weighted by information entropy, into the process of generating solutions, significantly enhancing convergence performance during the early-stage iterations. The combination of these strategies makes CDM-PSL have competitive performance in solving expensive multi-objective Bayesian optimization problem.

\section{Our Method}

% \textcolor{red}{one paragraph for overview.
% \\
% one figure framework\\
% one algorithm pseudocode\\}

% \subsection{MCTS}
% \subsection{CMD}
% \subsection{PtrNetwork?}

\begin{figure*}[ht]
\vskip 0.2in
\begin{center}
\centerline{\includegraphics[width=0.85\textwidth]{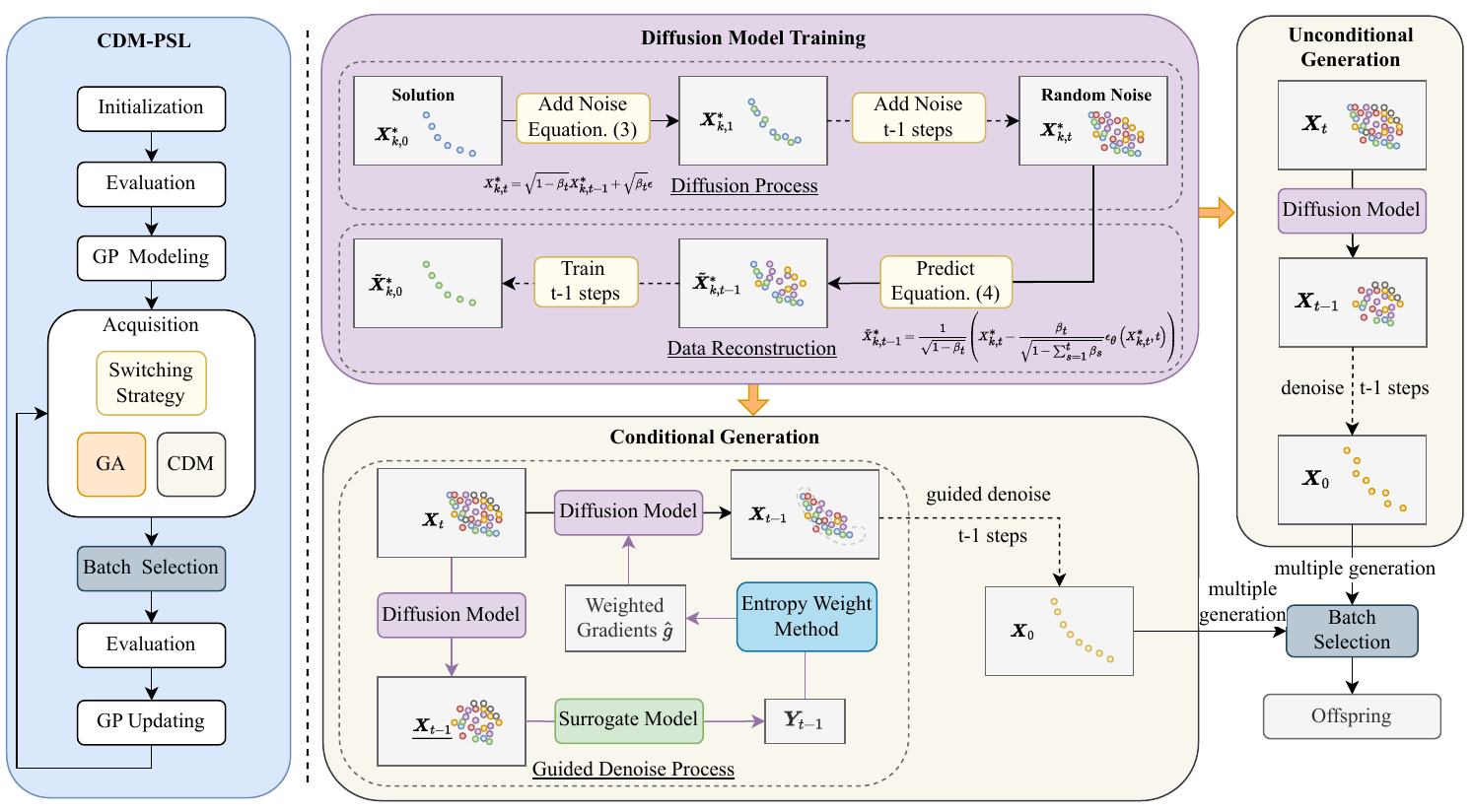}}
\caption{(left) The framework of CDM-PSL.
(right) Diffusion Model Training (DMT),
 Conditional Generation (CG),
 and Unconditional Generation (UG).
DMT involves learning from the selected samples through multiple steps; CG is designed to create 
high-quality samples with an optimized distribution;
UG is used to generate diverse samples with high efficiency.}
\label{flow}
\end{center}
\vskip -0.4in
\end{figure*}

\subsection{Overview}
We present a composite diffusion model based Pareto set learning method for EMOBO, denoted as CDM-PSL  (Algorithm \ref{cdmpsl} and Figure \ref{flow}). CDM-PSL contains three components to generate offspring: data extraction, diffusion model training and conditional generation. Initially, the process begins by initializing a set of samples $\boldsymbol{X}_0 \subset \mathcal{X}$, where $\mathcal{X} \subset \mathcal{R}^d$. These samples are drawn using \textit{Latin Hypercube Sampling} (LHS) \cite{mckay2000comparison}. Following this initialization, acquisition and batch selection are then applied in an iterative manner.

\begin{algorithm}[tb]
   \caption{The framework of MOBO with CDM-PSL}
   \label{cdmpsl}
\begin{algorithmic}
   \STATE {\bfseries Input:} black-box function $\boldsymbol{f}(\boldsymbol{x})$, number of iterations $K$, batch size $B$, number of initial solutions $N$\\
   \STATE {\bfseries Output:} final solutions   $\{\boldsymbol{X}_K,\boldsymbol{Y}_K\}$, Pareto front $\mathcal{P}_f$\\
   \STATE Initialize $N$ solutions $\{\boldsymbol{X}_0,\boldsymbol{Y}_0\}$ by LHS\\
   \FOR{$k=0$ {\bfseries to} $K-1$}
   \STATE Train surrogate model $GP_i^k$  based on $\{\boldsymbol{X}_k,\boldsymbol{Y}_k\}$ for each objective $f_i,i=1,\ldots,M$\\
   \STATE ${\boldsymbol{X}_k^{*}}$ $\leftarrow$ Data extraction based on $\{\boldsymbol{X}_k,\boldsymbol{Y}_k\}$ \textbf{(Algorithm \ref{de})}\\
   \IF{$\boldsymbol{F}_{CDM}$ is True}
   \STATE $\boldsymbol{S}$ $\leftarrow$ Pareto set learning based on ${\boldsymbol{X}_k^{*}}$ by CDM-PSL \textbf{(Algorithm \ref{psl})}\\
   \ELSE
   \STATE $\boldsymbol{S}$ $\leftarrow$ Generate offspring using \textit{Optimizer}\\
   \ENDIF
   \STATE ${\boldsymbol{X}}_k^B$ $\leftarrow$  Batch selection based on $GP^k$, $\boldsymbol{S}$ and $\{\boldsymbol{X}_k,\boldsymbol{Y}_k\}$\\
   \STATE Evaluate and update $\boldsymbol{X}_{k+1}$ $\leftarrow$ $\boldsymbol{X}_k$ $\cup$ ${\boldsymbol{X}_k^B}$, $\boldsymbol{Y}_{k+1}$ $\leftarrow$ $\boldsymbol{Y}_k$ $\cup$ $\boldsymbol{f}(\boldsymbol{X}_k^B)$\\
   \STATE Decide whether to invert $\boldsymbol{F}_{CDM}$ based on HV\\
   \ENDFOR
   \STATE Approximate the Pareto front $\mathcal{P}_f$ by non-dominated solutions in $\boldsymbol{Y}_K$
\end{algorithmic}
% \vspace{-0.2cm}
\end{algorithm}
\vspace{-0.2cm}

\begin{algorithm}[tb]
   \caption{Data Extraction}
   \label{de}
\begin{algorithmic}
   \STATE {\bfseries Input:} all already-evaluated solutions in the $k$-th iteration $\{\boldsymbol{X}_k,\boldsymbol{Y}_k\}$, number of real samples $T$\\
   \STATE {\bfseries Output:} solutions after data extraction   $\boldsymbol{X}_k^{*}$ \\
   \STATE $Fitness$ $\leftarrow$  Calculate the fitness of each solution in $\{\boldsymbol{X}_k,\boldsymbol{Y}_k\}$ by Eq. \ref{fit}\\
   \STATE $\boldsymbol{X}_k^{*}$ $\leftarrow$ Select top $T$ candidate solutions with $Fitness$\\
\end{algorithmic}
\end{algorithm}

\subsection{Data Extraction}
To prepare training data for Pareto set learning, we propose a data extraction strategy summarized in Algorithm \ref{de}. Central to this strategy is the application of shift-based density estimation (SDE) \cite{li2013shift} for calculating fitness values, which is mathematically represented as:
            \begin{eqnarray} 
                Fitness(\boldsymbol{p}) = \min \limits_{\boldsymbol{q} \in \boldsymbol{Y}_k \backslash \boldsymbol{p} }{\sqrt{\sum_{i=1}^{M}(
                \mathrm{max} 
                \{ 0,f_i(\boldsymbol{q})-f_i(\boldsymbol{p})
                \})^2}}
                \label{fit}
            \end{eqnarray} 
In this formula, $\boldsymbol{p}$ and $\boldsymbol{q}$ are solutions within the set $\boldsymbol{Y}_k$, and $f_i(\boldsymbol{p})$ indicates the $i$-th objective value of solution $\boldsymbol{p}$. The SDE methodology assesses the quality of samples based on their convergence and diversity characteristics. From $\boldsymbol{X}_k$, a total of $T$ candidate solutions (for instance, $\frac{|\boldsymbol{X}_k|}{3}$) that demonstrate superior SDE values are identified as Pareto optimal samples. This selection process is essential for ensuring the quality and relevance of the data used in Pareto set learning.

\begin{algorithm}[tb]
   \caption{Composite Diffusion Model based Generation}
   \label{psl}
\begin{algorithmic}
   \STATE {\bfseries Input:} trained model $\mathcal{M}$\\
   \STATE {\bfseries Parameter:} steps $T$, number of generation $N_1$, $N_2$\\
   \STATE {\bfseries Output:} solutions generated by CDM $\boldsymbol{S}$ \\
   \STATE $\boldsymbol{S}$ $\leftarrow$ $\emptyset$
   \FOR{$i=1$ {\bfseries to} $N_1$}
       \FOR{$t=T$ {\bfseries to} $1$}
       \STATE $z\sim\mathcal{N}(\mathbf{0},\mathbf{I})$\\
       \STATE $\hat{g}$ $\leftarrow$ Calculate weighted gradient by entropy weight method and surrogate model\\
       \STATE $x_{t-1}$ $\leftarrow$ $\frac1{\sqrt{\alpha_t}}(x_t-\frac{1-\alpha_t}{\sqrt{1-\bar{\alpha}_t}}\epsilon_\theta(x_t,t))+\sigma_t^2\hat{g}+\sigma_tz$\\
       \ENDFOR
       \STATE $\boldsymbol{S}$ $\leftarrow$ $\boldsymbol{S}$ $\cup$ $\{x_0\}$\\
   \ENDFOR
   \FOR{$i=1$ {\bfseries to} $N_2$}
       \FOR{$t=T$ {\bfseries to} $1$}
       \STATE $z\sim\mathcal{N}(\mathbf{0},\mathbf{I})$\\
       \STATE $x_{t-1}$ $\leftarrow$ $\frac1{\sqrt{\alpha_t}}(x_t-\frac{1-\alpha_t}{\sqrt{1-\bar{\alpha}_t}}\epsilon_\theta(x_t,t))+\sigma_tz$\\
       \ENDFOR
       \STATE $\boldsymbol{S}$ $\leftarrow$ $\boldsymbol{S}$ $\cup$ $\{x_0\}$\\
   \ENDFOR
\end{algorithmic}
\end{algorithm}

\subsection{Diffusion Model Training}
% In our pursuit to generate high-quality offspring, we introduce the diffusion model for Pareto Set Learning, with a focus on fitting the optimal samples.
The DM training process comprises two major steps: the diffusion process and noise prediction.

\subsubsection{Diffusion Process}
Given set of  samples $X_k^*$ and a specified step $t$, the diffusion process involves gradually introducing Gaussian noise $\epsilon \sim \mathcal{N}(0, I)$ to $X_k^*$ over $t$ steps:
	\begin{equation}
		X_{k,t}^*=\sqrt{1-\beta_t} X_{k,t-1}^* +\sqrt{\beta_t} \epsilon
		\label{addnoise}
	\end{equation}	
  % \vspace{-0.3cm}
In this equation, $X_{k,t}^*$ denotes the data at step $t$, and $\beta_t \in [1 \mathrm{e}-5, 5 \mathrm{e}-2]$ represents the noise level at step $t$. The DM learning process for PSL operates as a Markov chain. This stepwise approach simplifies the learning task compared to direct Pareto set learning and effectively captures the distribution characteristics of optimal samples.
\subsubsection{Noise Prediction}
The noise prediction phase involves reconstructing the samples $X_{k,t}^*$, which have undergone the diffusion process, back to their original state, $X_k^*$. This reconstruction is achieved through a model $\mathcal{M}$ that predicts the noise added at each step, thereby reversing the diffusion process. This process follows the equation \ref{predictnoise}:
	\begin{equation}
		\tilde{X}_{k,t-1}^*=\frac{1}{\sqrt{1-\beta_t}}\left(X_{k,t}^* -\frac{\beta_t}{\sqrt{1-\sum_{s=1}^t \beta_s}} \epsilon_\theta\left(X_{k,t}^*, t\right)\right).
		\label{predictnoise}
	\end{equation}
In this equation, $\tilde{X}_{k,t-1}^*$ represents the data after reconstruction, and $\theta$ denotes the parameters of the model. The term $\epsilon_\theta(X_{k,t}^*, t)$ is the predicted noise by model $\mathcal{M}$ at step $t$.\\
The loss function $\mathcal{L}$ for model $\mathcal{M}$ is defined as:
	\begin{equation}
		\mathcal{L}=\frac{1}{\mathcal{H}} \sum_{i=1}^\mathcal{H}\left(\epsilon-\epsilon_\theta\left(x_{k,i,t}^*, t\right)\right).
		\label{loss}
	\end{equation}
This equation takes $\mathcal{H}$ into account, the total number of optimal samples, where each $x_{k,i,t}^*$ is an instance from $X_{k,t}^*$, and $\epsilon_\theta(X_{k,t}^*, t)$ represents the predicted noise. The training of $\mathcal{M}$, using Equation \ref{loss}, aims to minimize the loss $\mathcal{L}$, thereby enhancing the accuracy of noise prediction.\\
The entire DM training process, comprising both the diffusion process and noise prediction, plays a crucial role in effective Pareto Set Learning for EMOBOPs. This method presents a novel approach to learning high-quality solutions, striking a balancing between exploration and exploitation in the search space.

\subsection{Conditional Generation}
In order to further enhancing the quality of generated samples, we propose a Guided Denoise Process. This process leverages gradients to guide the denoising procedure of DM, thereby achieving the realization of a Conditional Diffusion Model. The subsequent elucidation will provide a detailed explanation through two components: the guided denoise process and the weighted gradient.

\subsubsection{Guided Denoise Process}
For a given step $t$ and sample $x_t$, the denoising process with guidance can be implemented through equation \ref{guideddenoise}:
	\begin{equation}
		X_{t-1}=\frac1{\sqrt{\alpha_t}}(X_t-\frac{1-\alpha_t}{\sqrt{1-\bar{\alpha}_t}}\epsilon_\theta(X_t,t))+\sigma_t^2\hat{g}+\sigma_tz
		\label{guideddenoise}
	\end{equation}
where $\alpha_t$ represents $1-\beta_t$, $\epsilon_\theta(X_{t}, t)$ signifies the predicted noise by the trained model, $\sigma_t$ is the standard deviation of the $t$-th step, and $\hat{g}$ denotes the weighted gradients used to guide the denoising process.

To procure the gradients essential for guiding the model in sample generation, we establish separate Gaussian Process (GP) models for each objective, as proposed by Balandat \cite{balandat2020botorch}. These models are utilized to compute the objective values for all generated samples, thus obtaining gradients $g$ for each objective, facilitating the realization of conditional generation.

\subsubsection{Weighted Gradients}
In addressing EMOPs, employing weighted gradients is essential. This requires the determination of appropriate weights for each objective. Therefore, we propose a weighting methodology grounded in information entropy, facilitating the derivation of these weighted gradients $\hat{g}$.

In order to obtain weights based on information entropy, it is necessary to first normalize the objective values. Let $y_{ij}$ represent the $j$-th objective value of the $i$-th individual. The normalized objective value $\tilde{y}_{ij}$ is computed using Equation \ref{entropy1} as follows:
	\begin{equation}
		\tilde{y}_{ij}=\frac{y_{ij}-\min\left(y_{j}\right)}{\max(y_{j})-\min\left(y_{j}\right)}
		\label{entropy1}
	\end{equation}
 Subsequent to this, for each objective $j(j=1,2,\ldots,M)$, the probability matrix $P_{ij}$ is calculated using Equation \ref{entropy2}, where $k=1,2,\ldots,N$ and $N$ represents the number of individuals in the population.
 	\begin{equation}
		P_{ij}=\frac{\tilde{y}_{ij}}{\sum_{k=1}^{N}\tilde{y}_{kj}}
		\label{entropy2}
	\end{equation}
 Subsequently, for each objective $j(j=1,2,\ldots,M)$, the information entropy $E_j$ is computed. The method of calculation is as outlined in Equation \ref{entropy3}:
 	\begin{equation}
		E_{j}=-\frac{1}{\ln{(N)}}\sum_{i=1}^{N}\left(P_{ij}\times\ln(P_{ij}+\eta)\right)
		\label{entropy3}
	\end{equation}
Herein, $i=1,2,\ldots,N$ and $j=1,2,\ldots,M$, where $N$ signifies the total number of samples, and $M$ represents the quantity of objectives. To avoid the occurrence of $\ln(0)$, a small positive number, $\eta$, is introduced. The computation of information entropy relies on Shannon entropy \cite{shannon1948mathematical}, expressed as $H(X)=-\sum\left(P(x)\times\log\left(P(x)\right)\right)$. Additionally, the coefficient $\frac{1}{\ln\left(N\right)} $ is employed to guarantee that the values of information entropy are confined within the range of 0 to 1.

Finally, for each objective $j=1,2,\ldots,M$, the weight $W_j$ is computed using Equation \ref{entropy4}, where $k=1,2,\ldots,M$, and $M$ denotes the total number of objectives.
 	\begin{equation}
		W_{j}=\frac{1-E_{j}}{\sum_{k=1}^{M}(1-E_{k})}
		\label{entropy4}
	\end{equation}
By employing the calculated weights $W$, we can subsequently derive the entropy weighted gradients (EWG) $\hat{g}$. The entropy weighting method, serving as an adaptive weight allocation approach, assigns weights based on the information entropy of objectives. This method reduces subjectivity in weighting gradients for different objective values, and ensures that the algorithm places more emphasis on objectives with rich information content, thereby achieving better performance in multi-objective optimization problems. Furthermore, the characteristic of weight allocation based on information entropy endows the entropy weighting method with broad applicability, makeing it suitable for a wide range of fields and various types of multi-objective optimization problems, demonstrating its strong universality. In the context of guiding the denoising process, the weighted gradients also enhance the effectiveness of the guidance for multi-objective problems, thereby enhancing the conditional generation.

\subsection{Selection Strategy}
\subsubsection{Batch Selection}
After obtain the solutions sampled by the CDM-PSL, we employ the batch selection strategy of PSL-MOBO \cite{lin2022pareto} to select a small subset $\boldsymbol{X}_k^B=\{x_b|b=1,\ldots,B\}$. Specifically, this strategy use the Hypervolume (HV) indicator \cite{zitzler1999multiobjective} as the selection criteria. The HV indicator is defined as follows:
        \begin{eqnarray}
            \textbf{HV}(S) = \Lambda(\{q \in \mathbb{R}^d| \exists p \in S:p \leq q\; \textbf{and}\; q \leq r \})
            \label{hv}
        \end{eqnarray}
where $S$ signifies a solution set, $r$ is identified as a reference vector set, and $\Lambda(\cdot)$ denotes the Lebesgue measure. 
% The batch selection strategy is designed to maximize the HV improvement. Subsequently, the chosen subset $\boldsymbol{X}_k^B$ undergoes evaluation via true function evaluations (FE). The outcome, represented as $\{\boldsymbol{X}_k^B, \boldsymbol{Y}_k^B\}$, is then combined with the pre-existing set $\{\boldsymbol{X}_k, \boldsymbol{Y}_k\}$, culminating in the creation of the updated set $\{\boldsymbol{X}_{k+1}, \boldsymbol{Y}_{k+1}\}$.

\subsubsection{Operator Selection}
At the end of each iteration round, we have devised a method to switch operators based on the growth rate of the HV indicator. 
In algorithm 1, $F_{CDM}$ is a flag used to determine if CDM is currently being used to generate offspring.
After passing through an iteration window, such as three evaluation rounds, if the HV indicator's growth rate falls below a predefined threshold (default setting is 5\%), we switch the operator for offspring generation (including CDM-PSL and other \textit{Optimizer} like the Genetic Algorithm (GA)). This strategy is designed to prevent the algorithm from becoming trapped in local optima specific to the current operator. 
% The ablation study section validates the effectiveness of this approach.

\section{Experiments Study}

\begin{figure*}[h]
\vskip 0.2in
\begin{center}
\centerline{\includegraphics[width=0.86\textwidth]{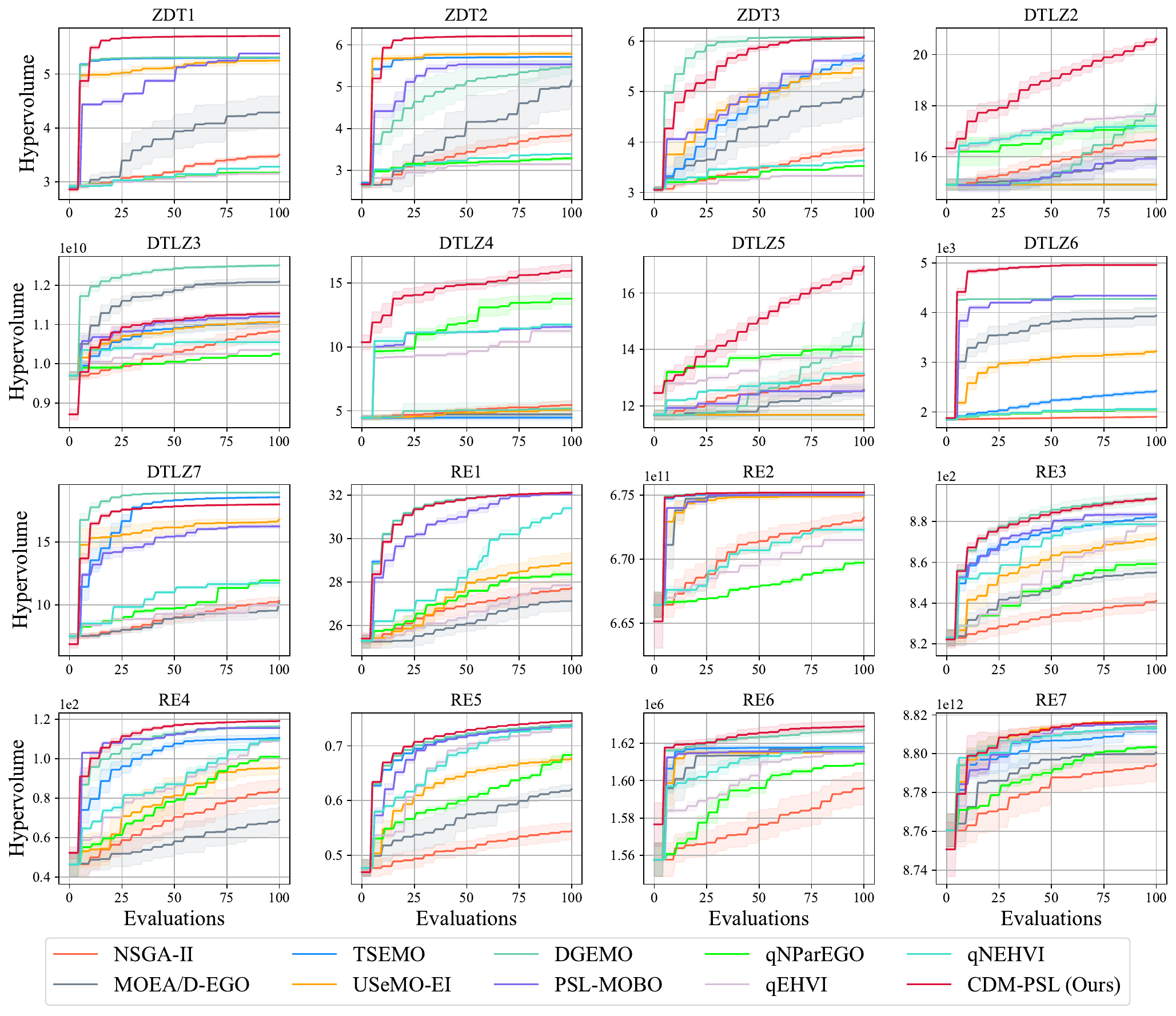}}
\caption{The HV results of 10 algorithms, evaluated on synthetic test functions and real-world problems ($d=20$).
The horizontal axis
denotes the FEs  after the initialization phase,  similarly hereinafter.}
\label{final_hv}
\end{center}
\vskip -0.4in
\end{figure*}

\begin{figure}[h]
\vskip 0.2in
\begin{center}
\centerline{\includegraphics[width=0.45\textwidth]{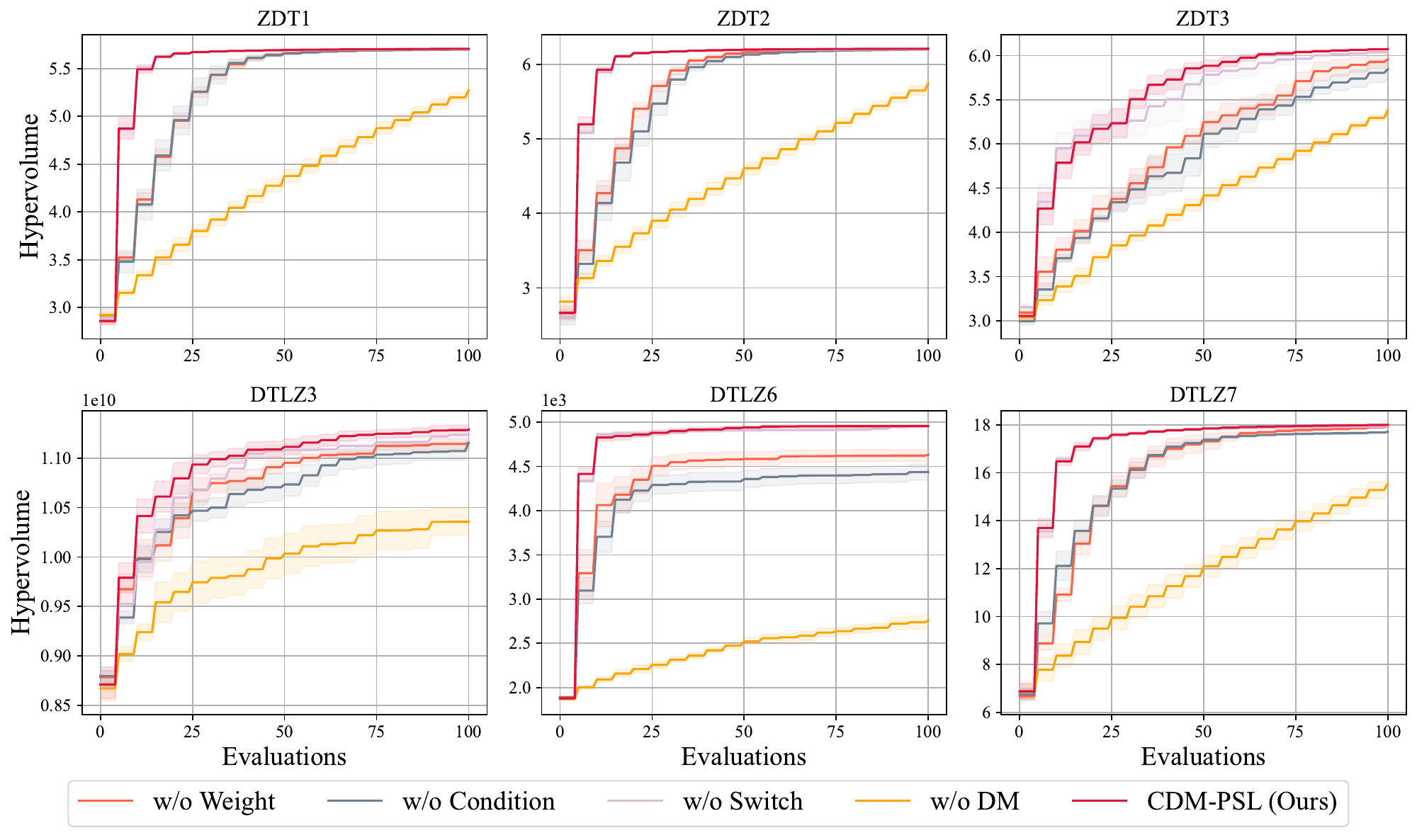}}
\caption{Ablation results of CDM-PSL on 6 instances.}
\label{ablation}
\end{center}
\vskip -0.4in
\end{figure}

\begin{figure}[h]
\vskip 0.2in
\begin{center}
% \centerline{\includegraphics[width=0.76\textwidth]{fig/ablation.pdf}}
    \subfigure[ZDT1]{\begin{minipage}[h]{0.15\textwidth}
        \includegraphics[width=1\textwidth]{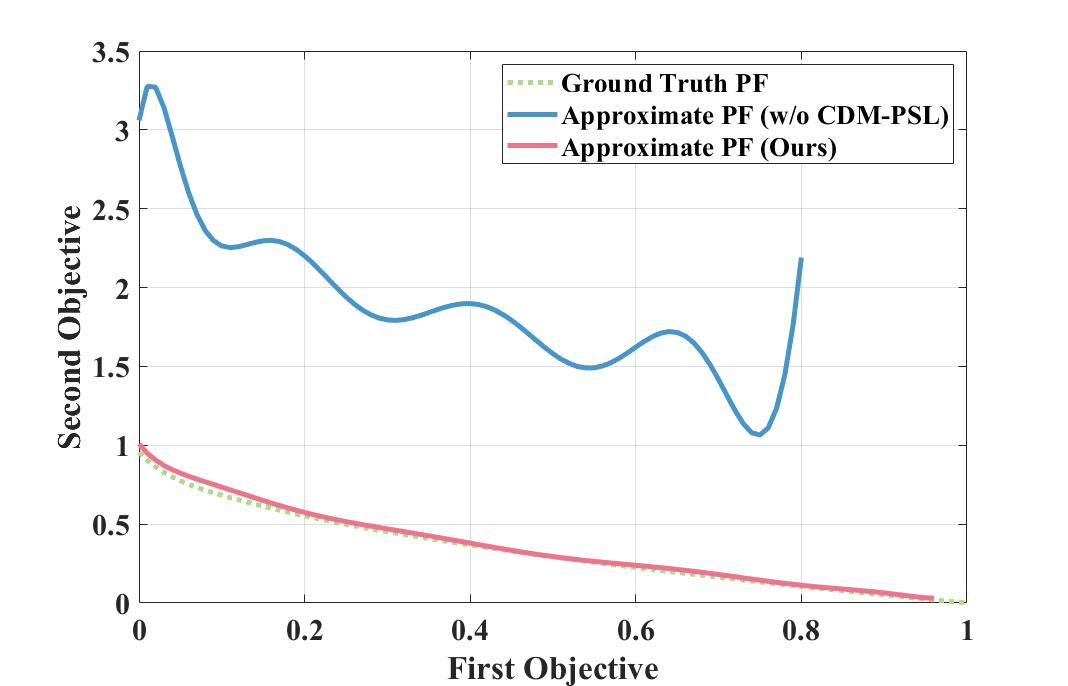}
    \end{minipage}}
    \subfigure[DTLZ6]{\begin{minipage}[h]{0.15\textwidth}
        \includegraphics[width=1\textwidth]{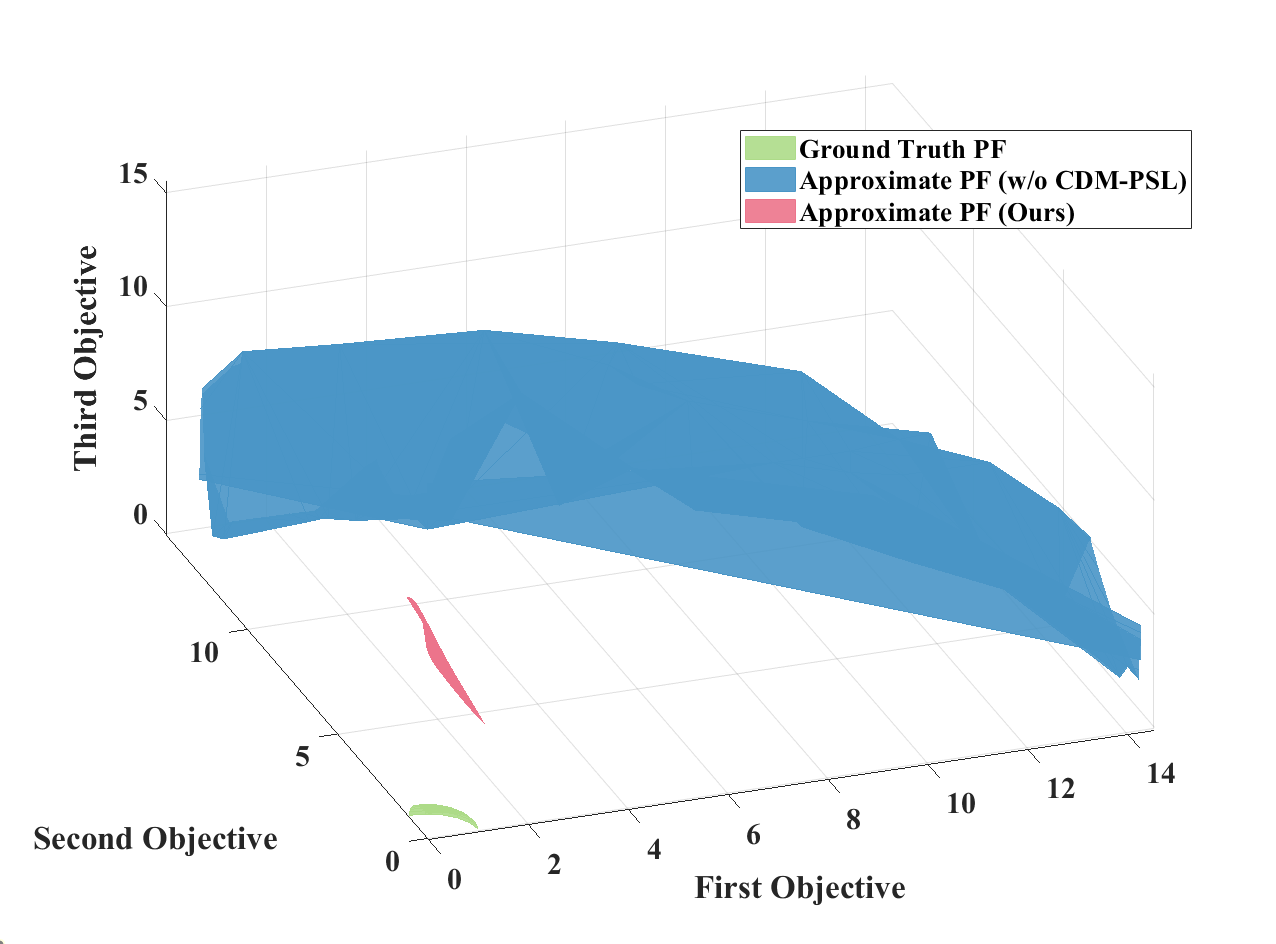}
    \end{minipage}}
    \subfigure[RE7]{\begin{minipage}[h]{0.15\textwidth}
        \includegraphics[width=1\textwidth]{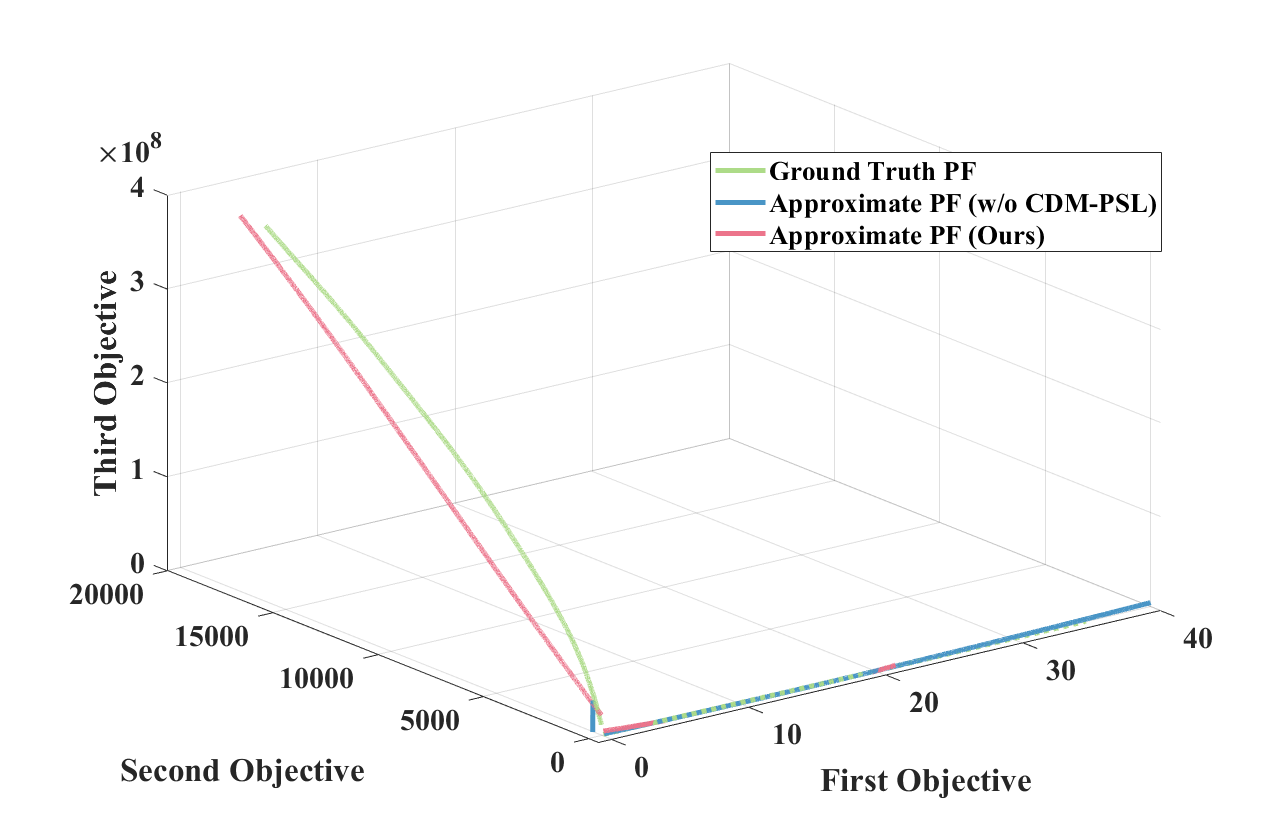}
    \end{minipage}}
\caption{Approximate Pareto fronts obtained by CDM-PSL and MOBO w/o CDM-PSL.}
\label{pf}
\end{center}
\vskip -0.5in
\end{figure}

\begin{figure}[h]
\vskip 0.2in
\begin{center}
\centerline{\includegraphics[width=0.46\textwidth]{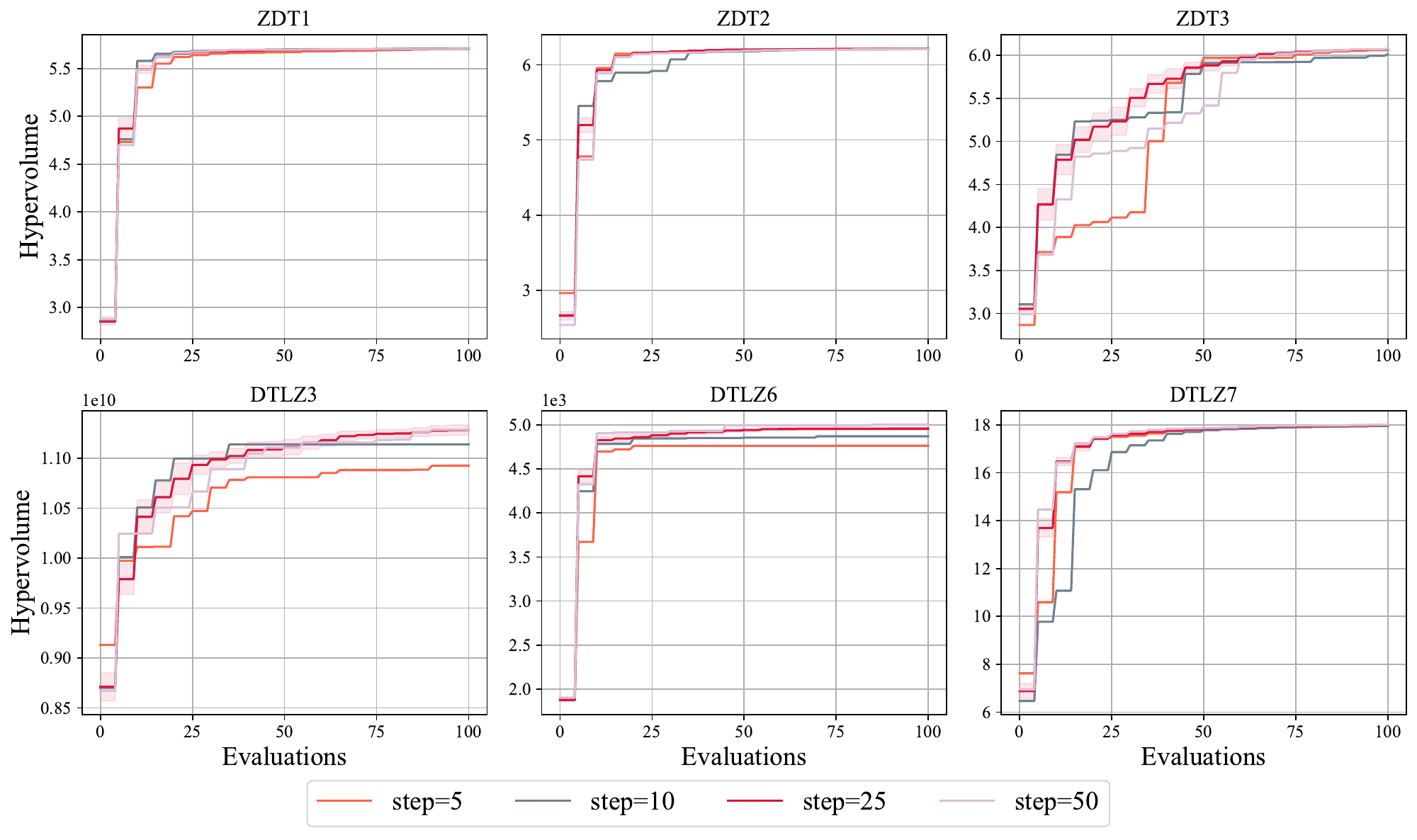}}
\caption{The HV values relative to the number of FEs for CDM-PSL with different number of steps $t$.}
\label{parameter}
\end{center}
\vskip -0.4in
\end{figure}

\subsection{Experimental Settings}
\paragraph{Instances and Baselines}
% \textcolor{magenta}{
To comprehensively validate the performance of CDM-PSL, experiments were conducted on 9 benchmark problems 
% \cite{zitzler2000comparison, deb2005scalable}
(2- and 3-objective
ZDT1-3 \cite{zitzler2000comparison} and DTLZ2-7 \cite{deb2005scalable})
and 7 real-world problems \cite{tanabe2020easy}. 
% To be specific,
% We carried out a series of experiments on 
% % a variety of widely recognized synthetic multi-objective benchmarks, 
% % including
% . 
% % The problems selected for the experiments featured
% % with 2 and 3 objectives.
% The number of decision variables ranged 
% from 10 to 20.} 

Moreover, 
we have compared CDM-PSL with 9 state-of-the-art and classical algorithms, including NSGA-II \cite{deb2002fast}, MOEA/D-EGO \cite{zhang2009expensive}, TSEMO \cite{bradford2018efficient}, USeMO-EI \cite{belakaria2020uncertainty}, DGEMO \cite{konakovic2020diversity}, PSL-MOBO \cite{lin2022pareto}, qNparEGO \cite{knowles2006parego}, qEHVI \cite{daulton2020differentiable} and qNEHVI \cite{daulton2021parallel}.
% \textcolor{red}{some sentences rough classfcications 
% and reasons of choosing
%  the compared algorithms}
 
\paragraph{Instances and Parameter Settings}
For fair comparison, the population size $N$ was initialized to 100 for all the compared algorithms. 
Bayesian optimization algorithms were executed for 20 batches, each with a batch size of 5, across all algorithms. 
Each method was randomly run 10 times. 
% \textcolor{red}{whether wilcoxon is used?
% The best (second-best) and its statistically insignificant
% results at 1/5\% significance level of a Wilcoxon rank-sum test are }
For CDM-PSL, the hyperparameter $t$ was set to 25, 
the batch size $m$ was 1024,
and the learning rate $\gamma$ was 0.001, with training spanning 4000 epochs. 
The configurations for other methods were aligned with those in their original publications (See the appendix). 
% (See Appendix\ref{appendixExperimentalSettings}).

\paragraph{Evaluation Metrics}
The hypervolume (HV) indicator, as defined in Equation \ref{hv}, was employed to assess the quality of the solutions obtained. 
% \textcolor{red}{why 200 not 100?}
Higher HV values are indicative of better performance.

% \begin{figure*}[ht]
% \vskip 0.2in
% \begin{center}
% \centerline{\includegraphics[width=0.86\textwidth]{fig/final_hv-mcts.pdf}}
% \caption{The HV values relative to the number of FEs for 10 algorithms, evaluated on synthetic test functions and real-world problems.}
% \label{final_hv}
% \end{center}
% \vskip -0.2in
% \end{figure*}

\subsection{Experimental Results}

We conducted a series of experiments on a variety of widely recognized synthetic multi-objective benchmarks, including ZDT1-3 \cite{zitzler2000comparison} and DTLZ2-7 \cite{deb2005scalable}. The problems selected for the experiments featured 2 and 3 objectives, with the number of decision variables set at 20. We particularly highlight the results for a specific instance where $d=20$, and further details are available in the supplementary materials. Additionally, the comprehensive performance
% \textcolor{red}{real-world can support scalability?\\}
of CDM-PSL was thoroughly evaluated through the application of seven real-world problems.

Figure \ref{final_hv} shows a comparison of the hypervolume (HV) indicator relative to function evaluations (FE). 
 CDM-PSL 
demonstrates outstanding performance across most synthetic benchmarks, excelling in both convergence speed and final values. 
Additionally, CDM-PSL exhibits ideal performance in real-world problems (RE). 
These findings decisively affirm the effectiveness and superiority of the CDM-PSL approach.

\subsection{Ablation Study}
To verify the effectiveness of each component in CDM-PSL, ablation study results are presented in Figure \ref{ablation}.

\paragraph{Entropy Weight}
CDM-PSL w/o Weight represents the CDM-PSL variant using mean weighted gradients to guide 
the sampling process instead of entropy weighted gradients (EWG).
CDM-PSL demonstrates superior convergence performance compared to CDM-PSL w/o Weight 
on all the six tested problems and attains better final values in ZDT3, DTLZ3 and DLTZ6.
This indicates that
\textcolor{black}{EWG}
can gauge the significance of objectives
more accurately, 
thereby offering more effective guidance for the sampling process.
\paragraph{Conditional Generation}
CDM-PSL w/o Condition refers to the variant from which the conditional generation component has been omitted. In a similar vein, CDM-PSL consistently and significantly outperforms CDM-PSL w/o Condition across all the tested problems. This clearly validates the critical role and effectiveness of the conditional generation component in the CDM-PSL.

\paragraph{Switching Strategy}
CDM-PSL w/o Switch is a variation of CDM-PSL that lacks the switching strategy. The results of the ablation experiments on ZDT3 and DTLZ3 demonstrate that the inclusion of this strategy prevents the PSL model from falling into local optima, achieved through alternating between various optimization operators. The incorporation of the switching strategy in CDM-PSL results in enhanced convergence performance.

\paragraph{Diffusion Model}
CDM-PSL w/o DM refers to the variant that employs the Genetic Algorithm (GA) instead of the DM-based PSL model as its optimizer, utilizing Simulated Binary Crossover (SBX) to generate new solutions. It is evident that this version's performance is markedly inferior compared to that of the standard CDM-PSL. This disparity underscores the effectiveness of DM based Pareto set learning.

\subsection{Validity of CDM-PSL}
Figure \ref{pf} displays the Pareto fronts approximated by CDM-PSL and by MOBO without CDM-PSL, based on the posterior mean. Clearly, CDM-PSL more effectively captures the essential features of the true PF, surpassing MOBO without CDM-PSL, in both synthetic benchmarks and real-world problems. For instance, MOBO without CDM-PSL is difficult to approach the true PF in a limited number of evaluations, yet CDM-PSL can effectively capture nearly all characteristics of ZDT1. In the three-objective DLTZ6 problem, the use of CDM-PSL can approximate the true PF faster in a limited number of evaluations. Additionally, our methodology demonstrates commendable exploitation capabilities on complex problems, such as the rocket injector design (RE7) \cite{vaidyanathan2003cfd}, which is characterized by a complex PF. This complexity arises from Pareto optimal solutions being distributed across multiple regions.

\subsection{Parameter Sensitivity Study}
The main parameter of CDM-PSL, step $t$, is studied in this subsection. Figure \ref{parameter} depicts the influence of varying $t$ values on CDM-PSL's performance. The default value of $t$ is set at 25 in CDM-PSL. A smaller $t$ yields suboptimal experimental outcomes; although near-excellent final values can be achieved on certain problems, there is a noticeable decline in convergence performance, attributed to the inadequate learning of the distribution of superior solutions. On the other hand, a larger $t$, such as 50, results in less stable convergence performance compared to $t=25$, primarily due to an increase in learning error. Furthermore, an increased step $t$ also leads to longer training durations. Consequently, balancing between convergence performance and time efficienvy, $t=25$ is established as the optimal default step size for CDM-PSL.

% \subsection{Results on Real-world Problems}
    
%     Table \ref{HV} shows the HV results of DGEA, FLEA, ALMOEA, GMOEA and PSL-DM on 5 real-world problems. It can be found that PSL-DM achieves the best results on  these problems. It should be pointed out that, the PSL-based methods, i.e. PSL-DM and GMOEA, demonstrate competitiveness on the 300 dimensional real-world problems.
    
%     \textcolor{black}{It should be noted that more details of the experimental results, including all outcomes of 10 methods (PSL-DM and 9 state-of-the-art MOEAs) across 40 benchmark problems and 5 real-word problems, a comparison of the running time for the 10 methods, and the results with different learning rates in PSL-DM, are provided in the supplementary material.}

\section{Conclusion and Future work}
In this paper, we introduce a composite diffusion model based Pareto set learning method, termed CDM-PSL, for addressing EMOBOPs.
CDM-PSL  uses both 
    unconditional and conditional diffusion model
    for generating high-quality samples.
    Besides, the quality of the solutions generated by the Pareto Set learning model is significantly enhanced by employing entropy weighted gradients to guide the sampling process. Extensive experimental evaluations on 9 benchmark problems and 7 real-world problems verify the efficiency of  CDM-PSL. 
% Notwithstanding its efficacy, CDM-PSL entails a complex training process, which leads to a relatively high computational overhead. Consequently, exploring methods to expedite the computational speed of CDM-PSL without compromising its performance constitutes an important direction for future research. 
The relatively high computational overhead makes CDM-PSL challenging to apply in very high-dimensional problems. The Monte Carlo tree approach has proven effective in feature extraction for such complex scenarios \cite{song2022monte}. Consequently, there is an intention to incorporate Monte Carlo trees into existing algorithms for future research to address higher-dimensional multi-objective Bayesian optimization problems.

\section*{Impact Statements}
This paper presents work whose goal is to advance the field of Multi-Objective Bayesian Optimization (MOBO). 
There are many potential societal consequences of our work, none which we feel must be specifically highlighted here.

\bibliography{icml2024/MCTS-CDM}

\begin{thebibliography}{69}
\providecommand{\natexlab}[1]{#1}
\providecommand{\url}[1]{\texttt{#1}}
\expandafter\ifx\csname urlstyle\endcsname\relax
  \providecommand{\doi}[1]{doi: #1}\else
  \providecommand{\doi}{doi: \begingroup \urlstyle{rm}\Url}\fi

\bibitem[Ament et~al.(2024)Ament, Daulton, Eriksson, Balandat, and Bakshy]{ament2024unexpected}
Ament, S., Daulton, S., Eriksson, D., Balandat, M., and Bakshy, E.
\newblock Unexpected improvements to expected improvement for bayesian optimization.
\newblock \emph{Advances in Neural Information Processing Systems}, 36, 2024.

\bibitem[Amir \& Hasegawa(1989)Amir and Hasegawa]{amir1989nonlinear}
Amir, H.~M. and Hasegawa, T.
\newblock Nonlinear mixed-discrete structural optimization.
\newblock \emph{Journal of Structural Engineering}, 115\penalty0 (3):\penalty0 626--646, 1989.

\bibitem[Austin et~al.(2021{\natexlab{a}})Austin, Johnson, Ho, Tarlow, and Van Den~Berg]{DM-NLP}
Austin, J., Johnson, D.~D., Ho, J., Tarlow, D., and Van Den~Berg, R.
\newblock Structured denoising diffusion models in discrete state-spaces.
\newblock \emph{Advances in Neural Information Processing Systems}, 34:\penalty0 17981--17993, 2021{\natexlab{a}}.

\bibitem[Austin et~al.(2021{\natexlab{b}})Austin, Johnson, Ho, Tarlow, and Van Den~Berg]{austin2021structured}
Austin, J., Johnson, D.~D., Ho, J., Tarlow, D., and Van Den~Berg, R.
\newblock Structured denoising diffusion models in discrete state-spaces.
\newblock \emph{Advances in Neural Information Processing Systems}, 34:\penalty0 17981--17993, 2021{\natexlab{b}}.

\bibitem[Balandat et~al.(2020)Balandat, Karrer, Jiang, Daulton, Letham, Wilson, and Bakshy]{balandat2020botorch}
Balandat, M., Karrer, B., Jiang, D., Daulton, S., Letham, B., Wilson, A.~G., and Bakshy, E.
\newblock Botorch: A framework for efficient monte-carlo bayesian optimization.
\newblock \emph{Advances in neural information processing systems}, 33:\penalty0 21524--21538, 2020.

\bibitem[Bao et~al.(2022)Bao, Li, Zhu, and Zhang]{bao2022analytic}
Bao, F., Li, C., Zhu, J., and Zhang, B.
\newblock Analytic-dpm: an analytic estimate of the optimal reverse variance in diffusion probabilistic models.
\newblock \emph{arXiv preprint arXiv:2201.06503}, 2022.

\bibitem[Baranchuk et~al.(2021)Baranchuk, Rubachev, Voynov, Khrulkov, and Babenko]{DM-CV1}
Baranchuk, D., Rubachev, I., Voynov, A., Khrulkov, V., and Babenko, A.
\newblock Label-efficient semantic segmentation with diffusion models.
\newblock \emph{arXiv preprint arXiv:2112.03126}, 2021.

\bibitem[Belakaria et~al.(2020)Belakaria, Deshwal, Jayakodi, and Doppa]{belakaria2020uncertainty}
Belakaria, S., Deshwal, A., Jayakodi, N.~K., and Doppa, J.~R.
\newblock Uncertainty-aware search framework for multi-objective bayesian optimization.
\newblock In \emph{Proceedings of the AAAI Conference on Artificial Intelligence}, volume~34, pp.\  10044--10052, 2020.

\bibitem[Bergstra et~al.(2011)Bergstra, Bardenet, Bengio, and K{\'e}gl]{bergstra2011algorithms}
Bergstra, J., Bardenet, R., Bengio, Y., and K{\'e}gl, B.
\newblock Algorithms for hyper-parameter optimization.
\newblock \emph{Advances in neural information processing systems}, 24, 2011.

\bibitem[Bradford et~al.(2018)Bradford, Schweidtmann, and Lapkin]{bradford2018efficient}
Bradford, E., Schweidtmann, A.~M., and Lapkin, A.
\newblock Efficient multiobjective optimization employing gaussian processes, spectral sampling and a genetic algorithm.
\newblock \emph{Journal of global optimization}, 71\penalty0 (2):\penalty0 407--438, 2018.

\bibitem[Cheng \& Li(1999)Cheng and Li]{cheng1999generalized}
Cheng, F. and Li, X.
\newblock Generalized center method for multiobjective engineering optimization.
\newblock \emph{Engineering Optimization}, 31\penalty0 (5):\penalty0 641--661, 1999.

\bibitem[Cornish et~al.(2020)Cornish, Caterini, Deligiannidis, and Doucet]{cornish2020relaxing}
Cornish, R., Caterini, A., Deligiannidis, G., and Doucet, A.
\newblock Relaxing bijectivity constraints with continuously indexed normalising flows.
\newblock In \emph{International conference on machine learning}, pp.\  2133--2143. PMLR, 2020.

\bibitem[Couckuyt et~al.(2014)Couckuyt, Deschrijver, and Dhaene]{couckuyt2014fast}
Couckuyt, I., Deschrijver, D., and Dhaene, T.
\newblock Fast calculation of multiobjective probability of improvement and expected improvement criteria for pareto optimization.
\newblock \emph{Journal of Global Optimization}, 60:\penalty0 575--594, 2014.

\bibitem[Daulton et~al.(2020)Daulton, Balandat, and Bakshy]{daulton2020differentiable}
Daulton, S., Balandat, M., and Bakshy, E.
\newblock Differentiable expected hypervolume improvement for parallel multi-objective bayesian optimization.
\newblock \emph{Advances in Neural Information Processing Systems}, 33:\penalty0 9851--9864, 2020.

\bibitem[Daulton et~al.(2021)Daulton, Balandat, and Bakshy]{daulton2021parallel}
Daulton, S., Balandat, M., and Bakshy, E.
\newblock Parallel bayesian optimization of multiple noisy objectives with expected hypervolume improvement.
\newblock \emph{Advances in Neural Information Processing Systems}, 34:\penalty0 2187--2200, 2021.

\bibitem[Deb \& Srinivasan(2006)Deb and Srinivasan]{deb2006innovization}
Deb, K. and Srinivasan, A.
\newblock Innovization: Innovating design principles through optimization.
\newblock In \emph{Proceedings of the 8th annual conference on Genetic and evolutionary computation}, pp.\  1629--1636, 2006.

\bibitem[Deb et~al.(2002)Deb, Pratap, Agarwal, and Meyarivan]{deb2002fast}
Deb, K., Pratap, A., Agarwal, S., and Meyarivan, T.
\newblock A fast and elitist multiobjective genetic algorithm: Nsga-ii.
\newblock \emph{IEEE transactions on evolutionary computation}, 6\penalty0 (2):\penalty0 182--197, 2002.

\bibitem[Deb et~al.(2005)Deb, Thiele, Laumanns, and Zitzler]{deb2005scalable}
Deb, K., Thiele, L., Laumanns, M., and Zitzler, E.
\newblock \emph{Scalable test problems for evolutionary multiobjective optimization}.
\newblock Springer, 2005.

\bibitem[Dhariwal \& Nichol(2021{\natexlab{a}})Dhariwal and Nichol]{DMbeatGAN}
Dhariwal, P. and Nichol, A.
\newblock Diffusion models beat gans on image synthesis.
\newblock \emph{Advances in Neural Information Processing Systems}, 34:\penalty0 8780--8794, 2021{\natexlab{a}}.

\bibitem[Dhariwal \& Nichol(2021{\natexlab{b}})Dhariwal and Nichol]{dhariwal2021diffusion}
Dhariwal, P. and Nichol, A.
\newblock Diffusion models beat gans on image synthesis.
\newblock \emph{Advances in neural information processing systems}, 34:\penalty0 8780--8794, 2021{\natexlab{b}}.

\bibitem[Ding et~al.(2019)Ding, Li, Li, and Yang]{ding2019compact}
Ding, D., Li, D., Li, Z., and Yang, L.
\newblock Compact circularly-polarized microstrip antenna for hand-held rfid reader.
\newblock In \emph{2019 8th Asia-Pacific Conference on Antennas and Propagation (APCAP)}, pp.\  181--182. IEEE, 2019.

\bibitem[Dung \& Binh(2022)Dung and Binh]{dung2022gdegan}
Dung, D.~A. and Binh, H. T.~T.
\newblock Gdegan: Graphical discriminative embedding gan for tabular data.
\newblock In \emph{2022 IEEE 9th International Conference on Data Science and Advanced Analytics (DSAA)}, pp.\  1--11. IEEE, 2022.

\bibitem[Fang et~al.(2024)Fang, Du, He, Tang, Jin, and Yen]{fang2024diffusion}
Fang, W., Du, W., He, R., Tang, Y., Jin, Y., and Yen, G.~G.
\newblock Diffusion model-based multiobjective optimization for gasoline blending scheduling.
\newblock \emph{arXiv preprint arXiv:2402.14600}, 2024.

\bibitem[Gong et~al.(2019)Gong, Xu, Li, Zhang, and Batmanghelich]{gong2019twin}
Gong, M., Xu, Y., Li, C., Zhang, K., and Batmanghelich, K.
\newblock Twin auxilary classifiers gan.
\newblock \emph{Advances in neural information processing systems}, 32, 2019.

\bibitem[Ho \& Salimans(2022)Ho and Salimans]{ho2022classifier}
Ho, J. and Salimans, T.
\newblock Classifier-free diffusion guidance.
\newblock \emph{arXiv preprint arXiv:2207.12598}, 2022.

\bibitem[Ho et~al.(2020)Ho, Jain, and Abbeel]{ho2020denoising}
Ho, J., Jain, A., and Abbeel, P.
\newblock Denoising diffusion probabilistic models.
\newblock \emph{Advances in neural information processing systems}, 33:\penalty0 6840--6851, 2020.

\bibitem[Ho et~al.(2022)Ho, Chan, Saharia, Whang, Gao, Gritsenko, Kingma, Poole, Norouzi, Fleet, et~al.]{ho2022imagen}
Ho, J., Chan, W., Saharia, C., Whang, J., Gao, R., Gritsenko, A., Kingma, D.~P., Poole, B., Norouzi, M., Fleet, D.~J., et~al.
\newblock Imagen video: High definition video generation with diffusion models.
\newblock \emph{arXiv preprint arXiv:2210.02303}, 2022.

\bibitem[Hoffman \& Ghahramani(2015)Hoffman and Ghahramani]{hoffman2015output}
Hoffman, M.~W. and Ghahramani, Z.
\newblock Output-space predictive entropy search for flexible global optimization.
\newblock In \emph{NIPS workshop on Bayesian Optimization}, pp.\  1--5, 2015.

\bibitem[Jin et~al.(2021)Jin, Netrapalli, Ge, Kakade, and Jordan]{jin2021nonconvex}
Jin, C., Netrapalli, P., Ge, R., Kakade, S.~M., and Jordan, M.~I.
\newblock On nonconvex optimization for machine learning: Gradients, stochasticity, and saddle points.
\newblock \emph{Journal of the ACM (JACM)}, 68\penalty0 (2):\penalty0 1--29, 2021.

\bibitem[Jones et~al.(1998)Jones, Schonlau, and Welch]{jones1998efficient}
Jones, D.~R., Schonlau, M., and Welch, W.~J.
\newblock Efficient global optimization of expensive black-box functions.
\newblock \emph{Journal of Global optimization}, 13\penalty0 (4):\penalty0 455, 1998.

\bibitem[Kang et~al.(2021)Kang, Shim, Cho, and Park]{kang2021rebooting}
Kang, M., Shim, W., Cho, M., and Park, J.
\newblock Rebooting acgan: Auxiliary classifier gans with stable training.
\newblock \emph{Advances in neural information processing systems}, 34:\penalty0 23505--23518, 2021.

\bibitem[Kannan \& Kramer(1994)Kannan and Kramer]{kannan1994augmented}
Kannan, B. and Kramer, S.~N.
\newblock An augmented lagrange multiplier based method for mixed integer discrete continuous optimization and its applications to mechanical design.
\newblock 1994.

\bibitem[Knowles(2006)]{knowles2006parego}
Knowles, J.
\newblock Parego: A hybrid algorithm with on-line landscape approximation for expensive multiobjective optimization problems.
\newblock \emph{IEEE Transactions on Evolutionary Computation}, 10\penalty0 (1):\penalty0 50--66, 2006.

\bibitem[Konakovic~Lukovic et~al.(2020)Konakovic~Lukovic, Tian, and Matusik]{konakovic2020diversity}
Konakovic~Lukovic, M., Tian, Y., and Matusik, W.
\newblock Diversity-guided multi-objective bayesian optimization with batch evaluations.
\newblock \emph{Advances in Neural Information Processing Systems}, 33:\penalty0 17708--17720, 2020.

\bibitem[Krishnamoorthy et~al.(2023)Krishnamoorthy, Mashkaria, and Grover]{krishnamoorthy2023diffusion}
Krishnamoorthy, S., Mashkaria, S.~M., and Grover, A.
\newblock Diffusion models for black-box optimization.
\newblock In \emph{International Conference on Machine Learning}, pp.\  17842--17857. PMLR, 2023.

\bibitem[Laumanns \& Ocenasek(2002)Laumanns and Ocenasek]{laumanns2002bayesian}
Laumanns, M. and Ocenasek, J.
\newblock Bayesian optimization algorithms for multi-objective optimization.
\newblock In \emph{Parallel Problem Solving from Nature—PPSN VII: 7th International Conference Granada, Spain, September 7--11, 2002 Proceedings 7}, pp.\  298--307. Springer, 2002.

\bibitem[Leng et~al.(2022)Leng, Chen, Guo, Liu, Chen, Tan, Mandic, He, Li, Qin, et~al.]{DM-Wave}
Leng, Y., Chen, Z., Guo, J., Liu, H., Chen, J., Tan, X., Mandic, D., He, L., Li, X., Qin, T., et~al.
\newblock Binauralgrad: A two-stage conditional diffusion probabilistic model for binaural audio synthesis.
\newblock \emph{Advances in Neural Information Processing Systems}, 35:\penalty0 23689--23700, 2022.

\bibitem[Letham et~al.(2019)Letham, Karrer, Ottoni, and Bakshy]{letham2019constrained}
Letham, B., Karrer, B., Ottoni, G., and Bakshy, E.
\newblock Constrained bayesian optimization with noisy experiments.
\newblock 2019.

\bibitem[Li et~al.(2013)Li, Yang, and Liu]{li2013shift}
Li, M., Yang, S., and Liu, X.
\newblock Shift-based density estimation for pareto-based algorithms in many-objective optimization.
\newblock \emph{IEEE Transactions on Evolutionary Computation}, 18\penalty0 (3):\penalty0 348--365, 2013.

\bibitem[Lin et~al.(2022)Lin, Yang, Zhang, and Zhang]{lin2022pareto}
Lin, X., Yang, Z., Zhang, X., and Zhang, Q.
\newblock Pareto set learning for expensive multi-objective optimization.
\newblock In \emph{36th Conference on Neural Information Processing Systems (NeurIPS 2022)}, 2022.

\bibitem[Liu et~al.(2022)Liu, Li, Ren, Chen, and Zhao]{liu2022diffsinger}
Liu, J., Li, C., Ren, Y., Chen, F., and Zhao, Z.
\newblock Diffsinger: Singing voice synthesis via shallow diffusion mechanism.
\newblock In \emph{Proceedings of the AAAI conference on artificial intelligence}, volume~36, pp.\  11020--11028, 2022.

\bibitem[Liu et~al.(2023)Liu, Park, Azadi, Zhang, Chopikyan, Hu, Shi, Rohrbach, and Darrell]{liu2023more}
Liu, X., Park, D.~H., Azadi, S., Zhang, G., Chopikyan, A., Hu, Y., Shi, H., Rohrbach, A., and Darrell, T.
\newblock More control for free! image synthesis with semantic diffusion guidance.
\newblock In \emph{Proceedings of the IEEE/CVF Winter Conference on Applications of Computer Vision}, pp.\  289--299, 2023.

\bibitem[Lu et~al.(2019)Lu, Whalen, Boddeti, Dhebar, Deb, Goodman, and Banzhaf]{lu2019nsga}
Lu, Z., Whalen, I., Boddeti, V., Dhebar, Y., Deb, K., Goodman, E., and Banzhaf, W.
\newblock Nsga-net: neural architecture search using multi-objective genetic algorithm.
\newblock In \emph{Proceedings of the genetic and evolutionary computation conference}, pp.\  419--427, 2019.

\bibitem[Lugmayr et~al.(2022)Lugmayr, Danelljan, Romero, Yu, Timofte, and Van~Gool]{lugmayr2022repaint}
Lugmayr, A., Danelljan, M., Romero, A., Yu, F., Timofte, R., and Van~Gool, L.
\newblock Repaint: Inpainting using denoising diffusion probabilistic models.
\newblock In \emph{Proceedings of the IEEE/CVF Conference on Computer Vision and Pattern Recognition}, pp.\  11461--11471, 2022.

\bibitem[McKay et~al.(2000)McKay, Beckman, and Conover]{mckay2000comparison}
McKay, M.~D., Beckman, R.~J., and Conover, W.~J.
\newblock A comparison of three methods for selecting values of input variables in the analysis of output from a computer code.
\newblock \emph{Technometrics}, 42\penalty0 (1):\penalty0 55--61, 2000.

\bibitem[Mo{\v{c}}kus(1975)]{movckus1975bayesian}
Mo{\v{c}}kus, J.
\newblock On bayesian methods for seeking the extremum.
\newblock In \emph{Optimization Techniques IFIP Technical Conference: Novosibirsk, July 1--7, 1974}, pp.\  400--404. Springer, 1975.

\bibitem[Odena et~al.(2017)Odena, Olah, and Shlens]{odena2017conditional}
Odena, A., Olah, C., and Shlens, J.
\newblock Conditional image synthesis with auxiliary classifier gans.
\newblock In \emph{International conference on machine learning}, pp.\  2642--2651. PMLR, 2017.

\bibitem[Preechakul et~al.(2022)Preechakul, Chatthee, Wizadwongsa, and Suwajanakorn]{preechakul2022diffusion}
Preechakul, K., Chatthee, N., Wizadwongsa, S., and Suwajanakorn, S.
\newblock Diffusion autoencoders: Toward a meaningful and decodable representation.
\newblock In \emph{Proceedings of the IEEE/CVF Conference on Computer Vision and Pattern Recognition}, pp.\  10619--10629, 2022.

\bibitem[Ray \& Liew(2002)Ray and Liew]{ray2002swarm}
Ray, T. and Liew, K.
\newblock A swarm metaphor for multiobjective design optimization.
\newblock \emph{Engineering optimization}, 34\penalty0 (2):\penalty0 141--153, 2002.

\bibitem[Shahriari et~al.(2015)Shahriari, Swersky, Wang, Adams, and De~Freitas]{shahriari2015taking}
Shahriari, B., Swersky, K., Wang, Z., Adams, R.~P., and De~Freitas, N.
\newblock Taking the human out of the loop: A review of bayesian optimization.
\newblock \emph{Proceedings of the IEEE}, 104\penalty0 (1):\penalty0 148--175, 2015.

\bibitem[Shannon(1948)]{shannon1948mathematical}
Shannon, C.~E.
\newblock A mathematical theory of communication.
\newblock \emph{The Bell system technical journal}, 27\penalty0 (3):\penalty0 379--423, 1948.

\bibitem[Snoek et~al.(2012)Snoek, Larochelle, and Adams]{snoek2012practical}
Snoek, J., Larochelle, H., and Adams, R.~P.
\newblock Practical bayesian optimization of machine learning algorithms.
\newblock \emph{Advances in neural information processing systems}, 25, 2012.

\bibitem[Sohl-Dickstein et~al.(2015)Sohl-Dickstein, Weiss, Maheswaranathan, and Ganguli]{sohl2015deep}
Sohl-Dickstein, J., Weiss, E., Maheswaranathan, N., and Ganguli, S.
\newblock Deep unsupervised learning using nonequilibrium thermodynamics.
\newblock In \emph{International conference on machine learning}, pp.\  2256--2265. PMLR, 2015.

\bibitem[Song et~al.(2022)Song, Xue, Huang, and Qian]{song2022monte}
Song, L., Xue, K., Huang, X., and Qian, C.
\newblock Monte carlo tree search based variable selection for high dimensional bayesian optimization.
\newblock \emph{Advances in Neural Information Processing Systems}, 35:\penalty0 28488--28501, 2022.

\bibitem[Song et~al.(2020)Song, Sohl-Dickstein, Kingma, Kumar, Ermon, and Poole]{song2020score}
Song, Y., Sohl-Dickstein, J., Kingma, D.~P., Kumar, A., Ermon, S., and Poole, B.
\newblock Score-based generative modeling through stochastic differential equations.
\newblock \emph{arXiv preprint arXiv:2011.13456}, 2020.

\bibitem[Tanabe \& Ishibuchi(2020)Tanabe and Ishibuchi]{tanabe2020easy}
Tanabe, R. and Ishibuchi, H.
\newblock An easy-to-use real-world multi-objective optimization problem suite.
\newblock \emph{Applied Soft Computing}, 89:\penalty0 106078, 2020.

\bibitem[Vaidyanathan et~al.(2003)Vaidyanathan, Tucker, Papila, and Shyy]{vaidyanathan2003cfd}
Vaidyanathan, R., Tucker, K., Papila, N., and Shyy, W.
\newblock Cfd-based design optimization for single element rocket injector.
\newblock In \emph{41st Aerospace sciences meeting and exhibit}, pp.\  296, 2003.

\bibitem[Wu et~al.(2020)Wu, K{\"o}hler, and No{\'e}]{wu2020stochastic}
Wu, H., K{\"o}hler, J., and No{\'e}, F.
\newblock Stochastic normalizing flows.
\newblock \emph{Advances in Neural Information Processing Systems}, 33:\penalty0 5933--5944, 2020.

\bibitem[Yan \& Jin(2024)Yan and Jin]{yan2024emodm}
Yan, X. and Jin, Y.
\newblock Emodm: A diffusion model for evolutionary multi-objective optimization.
\newblock \emph{arXiv preprint arXiv:2401.15931}, 2024.

\bibitem[Yang et~al.(2023)Yang, Zhang, Song, Hong, Xu, Zhao, Zhang, Cui, and Yang]{yang2023diffusion}
Yang, L., Zhang, Z., Song, Y., Hong, S., Xu, R., Zhao, Y., Zhang, W., Cui, B., and Yang, M.-H.
\newblock Diffusion models: A comprehensive survey of methods and applications.
\newblock \emph{ACM Computing Surveys}, 56\penalty0 (4):\penalty0 1--39, 2023.

\bibitem[Yu(1974)]{yu1974cone}
Yu, P.-L.
\newblock Cone convexity, cone extreme points, and nondominated solutions in decision problems with multiobjectives.
\newblock \emph{Journal of optimization Theory and Applications}, 14:\penalty0 319--377, 1974.

\bibitem[Yu et~al.(2022)Yu, Tack, Mo, Kim, Kim, Ha, and Shin]{DM-CV2}
Yu, S., Tack, J., Mo, S., Kim, H., Kim, J., Ha, J.-W., and Shin, J.
\newblock Generating videos with dynamics-aware implicit generative adversarial networks.
\newblock \emph{arXiv preprint arXiv:2202.10571}, 2022.

\bibitem[Yu et~al.(2019)Yu, Ramakrishnan, and Meinzer]{yu2019simulation}
Yu, Z., Ramakrishnan, V., and Meinzer, C.
\newblock Simulation optimization for bayesian multi-arm multi-stage clinical trial with binary endpoints.
\newblock \emph{Journal of Biopharmaceutical Statistics}, 29\penalty0 (2):\penalty0 306--317, 2019.

\bibitem[Zhang et~al.(2009)Zhang, Liu, Tsang, and Virginas]{zhang2009expensive}
Zhang, Q., Liu, W., Tsang, E., and Virginas, B.
\newblock Expensive multiobjective optimization by moea/d with gaussian process model.
\newblock \emph{IEEE Transactions on Evolutionary Computation}, 14\penalty0 (3):\penalty0 456--474, 2009.

\bibitem[Zhang et~al.(2015)Zhang, Sohn, Villegas, Pan, and Lee]{Zhang_2015_CVPR}
Zhang, Y., Sohn, K., Villegas, R., Pan, G., and Lee, H.
\newblock Improving object detection with deep convolutional networks via bayesian optimization and structured prediction.
\newblock In \emph{Proceedings of the IEEE Conference on Computer Vision and Pattern Recognition (CVPR)}, June 2015.

\bibitem[Zhao et~al.(2020)Zhao, Li, Yu, Gao, and Chen]{zhao2020feature}
Zhao, Y., Li, C., Yu, P., Gao, J., and Chen, C.
\newblock Feature quantization improves gan training.
\newblock \emph{arXiv preprint arXiv:2004.02088}, 2020.

\bibitem[Zitzler \& Thiele(1999)Zitzler and Thiele]{zitzler1999multiobjective}
Zitzler, E. and Thiele, L.
\newblock Multiobjective evolutionary algorithms: a comparative case study and the strength pareto approach.
\newblock \emph{IEEE transactions on Evolutionary Computation}, 3\penalty0 (4):\penalty0 257--271, 1999.

\bibitem[Zitzler et~al.(2000)Zitzler, Deb, and Thiele]{zitzler2000comparison}
Zitzler, E., Deb, K., and Thiele, L.
\newblock Comparison of multiobjective evolutionary algorithms: Empirical results.
\newblock \emph{Evolutionary computation}, 8\penalty0 (2):\penalty0 173--195, 2000.

\bibitem[Zoph et~al.(2018)Zoph, Vasudevan, Shlens, and Le]{zoph2018learning}
Zoph, B., Vasudevan, V., Shlens, J., and Le, Q.~V.
\newblock Learning transferable architectures for scalable image recognition.
\newblock In \emph{Proceedings of the IEEE conference on computer vision and pattern recognition}, pp.\  8697--8710, 2018.

\end{thebibliography}
\bibliographystyle{icml2024}

%%%%%%%%%%%%%%%%%%%%%%%%%%%%%%%%%%%%%%%%%%%%%%%%%%%%%%%%%%%%%%%%%%%%%%%%%%%%%%%
%%%%%%%%%%%%%%%%%%%%%%%%%%%%%%%%%%%%%%%%%%%%%%%%%%%%%%%%%%%%%%%%%%%%%%%%%%%%%%%
% APPENDIX
%%%%%%%%%%%%%%%%%%%%%%%%%%%%%%%%%%%%%%%%%%%%%%%%%%%%%%%%%%%%%%%%%%%%%%%%%%%%%%%
%%%%%%%%%%%%%%%%%%%%%%%%%%%%%%%%%%%%%%%%%%%%%%%%%%%%%%%%%%%%%%%%%%%%%%%%%%%%%%%
\newpage
\appendix
\onecolumn
% \section{Detailed Experimental Settings}
% \label{appendixExperimentalSettings}
% \section{Details of Our Proposed xxx}
% \subsection{Model Structure or parameters}
% you can 
% \section{Detailed Experimental Results}
% \subsection{Results on XXX Problems}
% \subsection{Results on Real-world Problems}

% \subsection{Parameter Sensitivity Analysis}
% \subsection{You \emph{can} have an appendix here.}

% You can have as much text here as you want. The main body must be at most $8$ pages long.
% For the final version, one more page can be added.
% If you want, you can use an appendix like this one.  

% The $\mathtt{\backslash onecolumn}$ command above can be kept in place if you prefer a one-column appendix, or can be removed if you prefer a two-column appendix.  Apart from this possible change, the style (font size, spacing, margins, page numbering, etc.) should be kept the same as the main body.

\section{Experimental Details}
\subsection{Parameter Settings}

First, we supplement more details about training DM. DM is designed with a straightforward architecture, incorporating two linear layers, each containing 128 hidden units, and utilizes the ReLU activation function. Additionally, we train DM for 4000 epochs, with a batch size of 1024. The Adam optimizer is selected for the model, set with a learning rate of 0.0001.

Moreover, the hyperparameter for the DM’s step t is configured at 25, and the noise level is defined within the range from 1e-5 to 0.5e-1.

The maximum number of function evaluations is 200, where 100 FEs are used to initialize the population and 100 FEs are used to evaluate new offspring after batch selection. Besides, batch size of all the algorithms is 5, thus number of iterations is 100 / 5 = 20.

In CDM-PSL, the initial populations are all obtained from the Latin Hypercube Sampling (LHS) \cite{mckay2000comparison}. To ensure fairness and allow the comparison algorithms to perform as them originally did, the relevant settings of the comparison algorithms are kept unchanged from the original paper.

\subsection{Synthetic Benchmark Problems}
In this subsection, we detail each synthetic benchmark problem, outlining the dimensions of the decision space $\mathcal{X} \in \mathbb{R}^d$ and the objective space $\boldsymbol{f}(\mathcal{X}) \in \mathbb{R}^M$, where $\boldsymbol{f}(\cdot)$ represents the black-box function. Additionally, we discuss the reference points $\boldsymbol{r}$ utilized for computing the hypervolume indicator. Our experiments encompass 9 synthetic benchmark problems, with the objectives numbering 2 and 3, and the decision variables is 10, 20 and 50. Further details are available in Table \ref{synthetic}, and the characteristics of each problem are listed in Table \ref{characteristics}. For the reference point $\boldsymbol{r} \in \mathbb{R}^M$, we adopt a vector comprising the maximum objective values from the initial solution set ${\boldsymbol{x}_1,...,\boldsymbol{x}_N}$ (Eq. \ref{mmax}).
\begin{eqnarray} 
\begin{aligned}
        &\boldsymbol{r} = (max_{1 \leq i \leq N}{\boldsymbol{f}_1(\boldsymbol{x}_i)},...,max_{1 \leq i \leq N}{\boldsymbol{f}_M(\boldsymbol{x}_i)})
        \label{mmax}
\end{aligned}        
\end{eqnarray}
\begin{table}[!h]  
    %  \scriptsize
    
    \footnotesize
        % \tiny\scriptsize\footnotesize\small\normalsize\large\Large\LARGE\huge\Huge
 % \tabcolsep0.01in
 \renewcommand{\arraystretch}{1.3}
  \caption{Description of synthetic benchmarks we used in this work.}
  \label{synthetic}
    \centering
  \begin{tabular}{lllll}
    \toprule
    Problem&M&d&reference point ($\boldsymbol{r}$)\\
    \midrule
    ZDT1 &10&2 &(0.9901, 5.8380)\\
    &20&2&(0.9994, 6.0576)\\
    &50&2&(0.9926, 5.8722)\\
    \midrule
    ZDT2 &10&2 &(0.9901, 7.4625)\\
    &20&2&(0.9994, 6.8960)\\
    &50&2&(0.9926, 6.2181)\\
    \midrule
    ZDT3 &10&2 &(0.9901, 5.8101)\\
    &20&2&(0.9994, 6.0571)\\
    &50&2&(0.9926, 5.8648)\\
    \midrule
    DTLZ2 &10&3 &(1.8753, 1.9635, 2.0503)\\
    &20&3&(2.8390, 2.9011, 2.8575)\\
    &50&3&(5.4059, 5.6415, 6.5147)\\
    \midrule
    DTLZ3 &10&3 &(1002.2772, 1173.7753, 1075.0803)\\
    &20&3&(2421.6427, 1905.2767, 2532.9691)\\
    &50&3&(5298.0203, 5548.5643, 6250.5483)\\
    \midrule
    DTLZ4 &10&3 &(2.1527, 1.2691, 0.9761)\\
    &20&3&(3.2675, 2.6443, 2.4263)\\
    &50&3&(6.9024, 4.2313, 3.61552)\\
    \midrule
    DTLZ5 &10&3 &(1.7079, 1.7896, 2.0503)\\
    &20&3&(2.6672, 2.8009, 2.8575)\\
    &50&3&(5.2481, 5.5483, 6.5147)\\
    \midrule
    DTLZ6 &10&3 &(8.1417, 8.0853, 8.5711)\\
    &20&3&(16.8258, 16.9194, 17.7646)\\
    &50&3&(42.4261, 44.4857, 45.2363)\\
    \midrule
    DTLZ7 &10&3 &(0.9902, 0.9932, 24.5440)\\
    &20&3&(0.9984, 0.9961, 22.8114)\\
    &50&3&(0.9925, 0.9954, 21.7584)\\
    \bottomrule
    
  \end{tabular}
\end{table}
\begin{table}[htbp]
        % \footnotesize
        % \Large
        % \tiny\scriptsize\footnotesize\small\normalsize\large\Large\LARGE\huge\Huge
			\begin{center}
				\caption{Characteristics of the synthetic benchmark problems.}
				\label{characteristics}
				\begin{tabular}{c|c|c|c}
					
					\toprule
					Benchmark &Features of&Features&Separability of\\
					problem&Pareto front&of landscape&objective functions\\
					
					\midrule
					DTLZ1& Linear& Multimodal &Fully Separable\\
					DTLZ2 &Concave& Unimodal& Fully Separable\\
					
					DTLZ3 &Concave& Multimodal& Fully Separable\\
					
					DTLZ4& Concave& Unimodal, biased& Fully Separable\\
					
					DTLZ5& Concave, degenerate& Unimodal &Fully Separable\\
					
					DTLZ6 &Concave, degenerate &Unimodal, biased &Fully Separable\\
					
					DTLZ7& Disconnected& Multimodal& Fully Separable\\

					\bottomrule
					
				\end{tabular}
			\end{center}
		\end{table}
  
\subsection{Real-world Application Problems}
Our experiments include 7 real-world application problems \cite{tanabe2020easy}, alongside several synthetic benchmark problems. These problems were originally proposed across various fields for distinct applications. Below are introductions to each of these real-world problems:

\paragraph{Four Bar Truss Design (RE1).} This task aims to optimize the design of a four-bar truss, focusing on minimizing structural volume ($f_1$) and joint displacement ($f_2$). It considers the lengths of the four bars ($x_1$, $x_2$, $x_3$, and $x_4$) as variables. For more information, please see \cite{cheng1999generalized}.

\paragraph{Pressure Vessel Design (RE2).} The goal here is to design a cylindrical pressure vessel to minimize total costs ($f_1$), including materials, forming, and welding, and to avoid violations of three design constraints ($f_2$). The decision variables are the shell thicknesses ($x_1$), the pressure vessel head ($x_2$), the inner radius ($x_3$), and the length of the cylindrical section ($x_4$). Additional details are in \cite{kannan1994augmented}.

\paragraph{Disk Brake Design (RE3).} This problem focuses on designing a disc brake to minimize mass ($f_1$), stopping time ($f_2$), and violations of four design constraints ($f_3$), with four decision variables: inner radius ($x_1$), outer radius ($x_2$), engaging force ($x_3$), and the number of friction surfaces ($x_4$). Further details are available in \cite{ray2002swarm}.

\paragraph{Gear Train Design (RE4).} The objective is to design a gear train to minimize the deviation from the required gear ratio ($f_1$), the maximum size of the gears ($f_2$), and violations of design constraints ($f_3$), considering the number of teeth in each of the four gears ($x_1$, $x_2$, $x_3$, and $x_4$). More information can be found in \cite{deb2006innovization}.

\paragraph{Rocket Injector Design (RE5).} The design goal for the rocket injector is to minimize the maximum temperature on the injector face ($f_1$), the distance from the inlet ($f_2$), and the temperature at the post tip ($f_3$), with decision variables including hydrogen flow angle ($x_1$), hydrogen area ($x_2$), oxygen area ($x_3$), and oxidizer post tip thickness ($x_4$). Detailed information is in \cite{vaidyanathan2003cfd}.

\paragraph{Reinforced Concrete Beam Design (RE6).} This design problem for a reinforced concrete beam aims to minimize the total cost of concrete and reinforcing steel ($f_1$) and the sum of two constraint violations ($f_2$), with variables for the reinforcement area ($x_1$), beam width ($x_2$), and beam depth ($x_3$). Details are available in \cite{amir1989nonlinear}.

\paragraph{Welded Beam Design (RE7).}  The challenge is to design a welded beam to minimize the cost ($f_1$), end deflection ($f_2$), and violations of four constraints ($f_3$), with variables adjusting the beam's size ($x_1$, $x_2$, $x_3$, and $x_4$). More information can be found in \cite{ray2002swarm}.

To maintain consistency in evaluation, the same reference point was applied when evaluating all the  algorithms.
More details are shown in the supplementary material. 
Details on the real-world application problems, including information and reference points, are presented in Table \ref{real}.
\begin{table}[htbp]  
    %  \scriptsize
    
    \small
        % \tiny\scriptsize\footnotesize\small\normalsize\large\Large\LARGE\huge\Huge
 % \tabcolsep0.01in
 \renewcommand{\arraystretch}{1.3}
  \caption{Description of real-world application problems we used in this work.}
  \label{real}
    \centering
  \begin{tabular}{lllcl}
    \toprule
    Problem&d&M&reference point ($\boldsymbol{r}$)\\
    \midrule
    RE1 &4&2 &(2763.2229, 0.0369)\\
    RE2 &4&2 &(528107.1899, 1279320.8107)\\
    RE3 &4&3 &(7.6853, 7.2861, 21.5010)\\
    RE4 &4&3 &(6.7921, 60.0, 0.48)\\
    RE5 &4&3 &(0.8745, 1.05092, 1.0533)\\
    RE6 &3&2 &(749.9241, 2229.3748)\\
    RE7 &4&3 &(210.3363, 1069.9160, 39196770.1704)\\
    \bottomrule
    
  \end{tabular}
\end{table}

\subsection{Hardware Settings}
The experiments in this paper are conducted on a server with Ubuntu 22.04 LTS operating system, a 1.5GHz AMD EPYC 7742 CPU (64 CPU Cores), 320 GB RAM, and two NVIDIA RTX 4090 GPU.

\section{Related Works}

\subsection{Comparing DM-Based Methods with Normalizing Flows}

Normalizing flows (NF) transforms data into a prior distribution through bijective functions, which limits the ability of modeling complex data distribution in both practical and theoretical contexts \cite{cornish2020relaxing, wu2020stochastic}. However, DM is more efficient in the task of \textbf{capturing complex data distributions}, and the CDM-PSL proposed in this paper also demonstrates this experimentally. Moreover, the random noise introduced by DM can also naturally \textbf{generates a greater diversity of candidate solutions}.

Specifically, DM enhances its expressive power by introducing random noise in the forward and backward processes, compared with NF. Considering the case of generating candidate solutions, this property of DM enables the exploration of more solutions in the space, thus enriching the diversity of solutions.

From the perspective of the motivation behind our proposed method, the performance of PSL methods might deteriorate significantly due to the complexity of distributions of Pareto optimal solutions and the limitation of function evaluations in the face of EMOBOPs. \textbf{DM can model complex distributions of solutions step by step, and generate more diverse solutions by introducing random noise}. This gives DM an advantage over NF for application on the PSL of EMOBOPs.

\subsection{Comparing CDM-PSL with DDOM}

In this section, we explain the novelty and advantages of CDM over DDOM. Krishnamoorthy proposed an offline black-box optimization algorithm named DDOM \cite{krishnamoorthy2023diffusion}, based on the diffusion model. DDOM trains a generative model on offline dataset and proof its effectiveness on single-objective optimization problems. The novelty and advantages of CDM-PSL over DDOM are as follows:

\textbf{Model composition:} The main difference between CDM-PSL and DDOM is that CDM consists of two generative models, CG and UG. CG can minimize all the objective values as much as possible in a limited number of function evaluations, and UG can generate solutions with greater diversity.

\textbf{Training data:} CDM-PSL trains DM on online dataset, while DDOM is trained on offline dataset, which means that CDMs are able to receive and learn new data instantly, which allows them to quickly adapt to changes in data distribution.

\textbf{Inference process:} CDM-PSL introduces gradient information, weighted by information entropy, into the process of generating solutions, significantly enhancing convergence performance under limited FEs.

\textbf{Research problems:} CDM-PSL is targeted at solving multi-objective optimization problems, while DDOM mainly solves single-objective optimization problems.

The combination of our proposed methods makes CDM-PSL have competitive performance in solving EMOPs.

\section{Additional Experiments}

\subsection{The Advantages of Using Diffusion Model}

\begin{figure*}[h]
\vskip 0.2in
\begin{center}
% Adjust the widths as necessary to fit side by side
\includegraphics[width=0.50\textwidth]{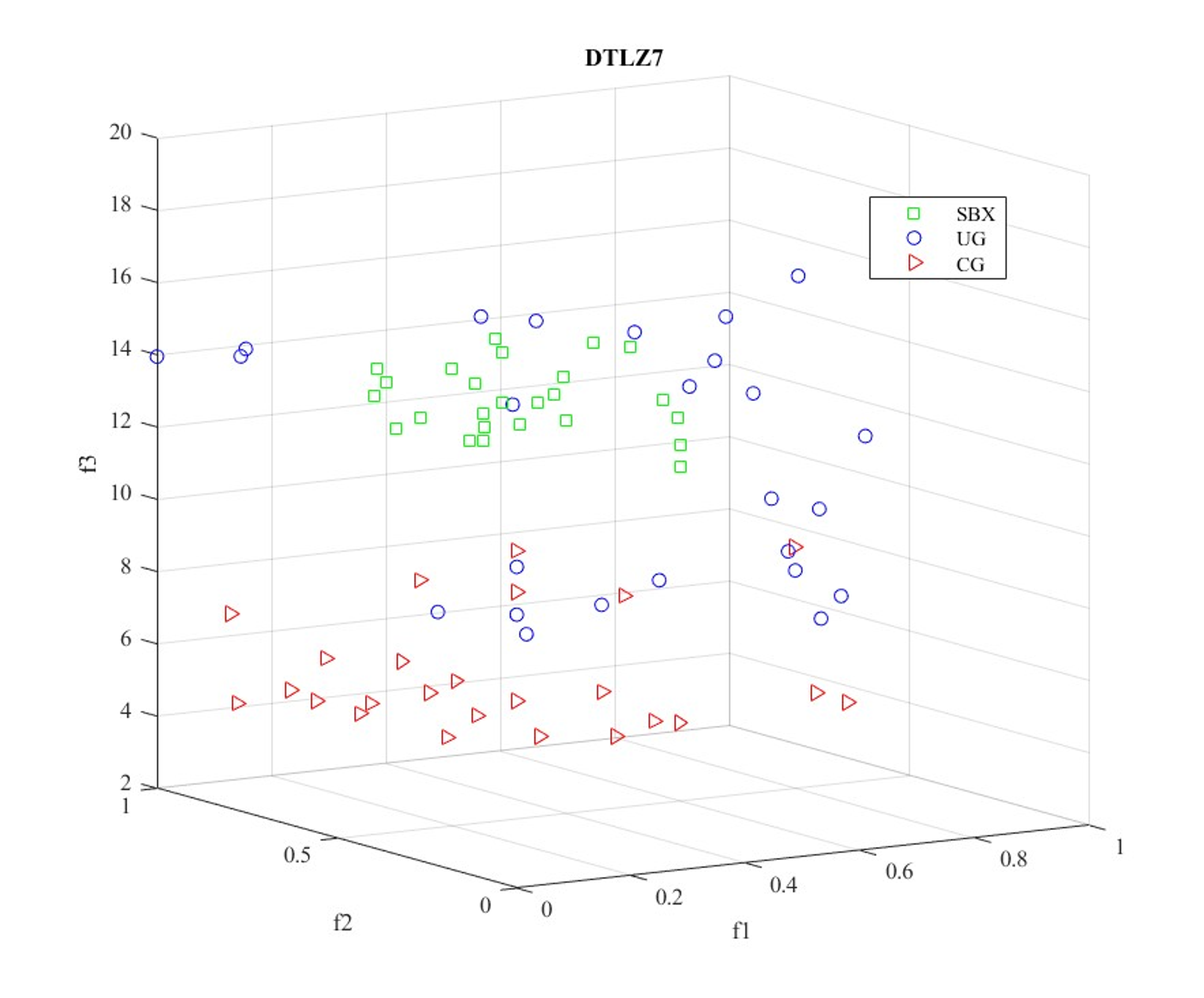}
\includegraphics[width=0.48\textwidth]{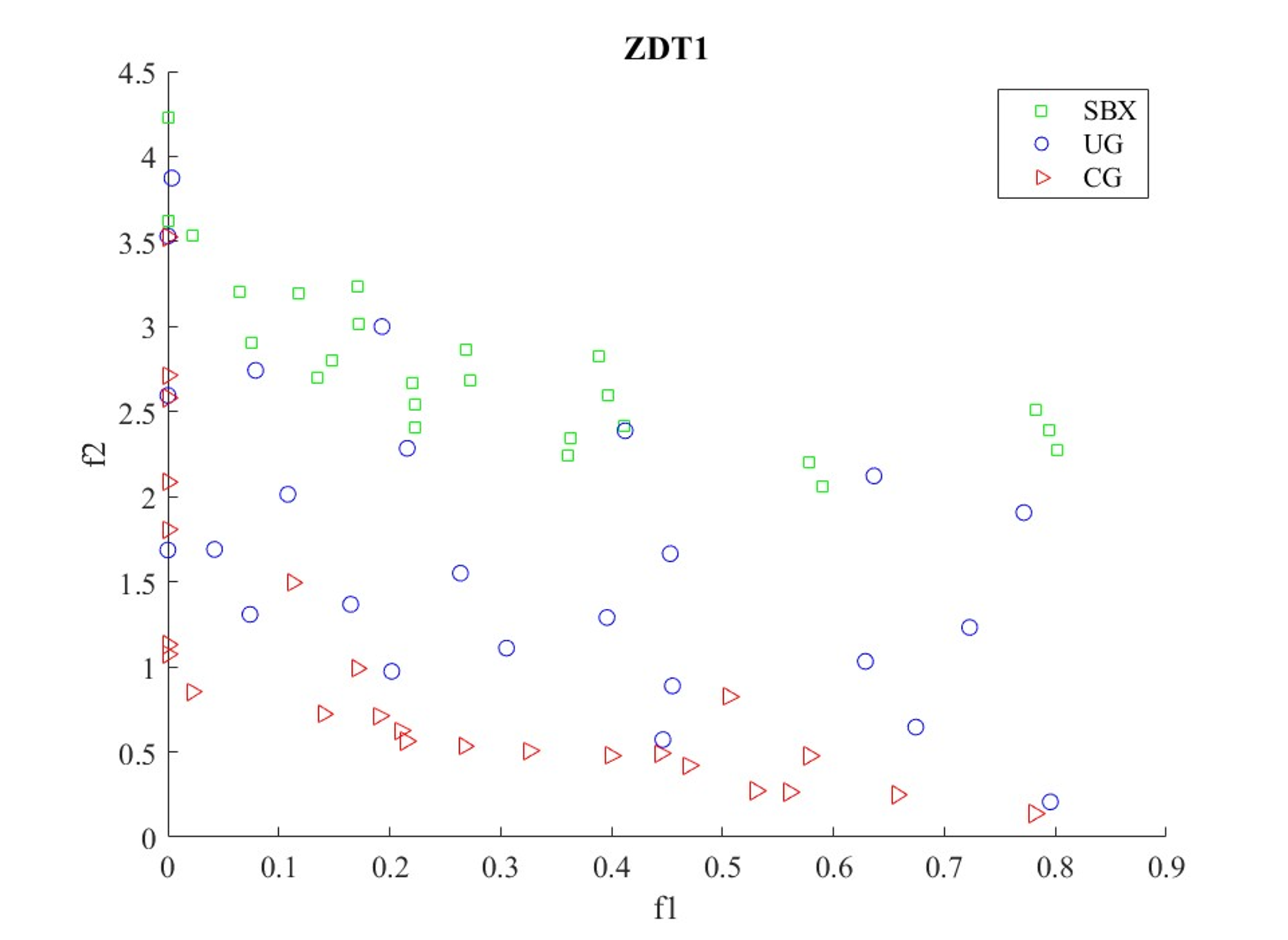}
\caption{The objective values of newly generated solutions obtained by GA's SBX method, UG and CG. ($d = 20$, $FE = 25$)}
\label{sbx-ug-cg}
\end{center}
\vskip -0.2in
\end{figure*}

Effectively modeling the complicated distribution of the PS is quite challenging for EMOPs with limited function evaluations. Existing Pareto set learning algorithms may exhibit considerable instability in expensive scenarios \cite{lin2022pareto}. DM is meticulously engineered to represent complex datasets through a highly flexible family of probability distributions. It has demonstrated promising outcomes in several domains, notably in image restoration and data synthesis \cite{sohl2015deep}. Thus, DM offers a promising way for Pareto set learning in expensive multi-objective Bayesian optimization.
Specifically, the proposed CDM-PSL is a composite of CG with conditional DM and UG with unconditional DM. It takes advantage of \textbf{CG's convergence capabilities with UG's diversity potential}. Figure \ref{sbx-ug-cg} illustrates the offspring produced by the SBX versus those generated by CG and UG on DTLZ7 and ZDT1 at FE = 25. The closer to (0, 0, 0), the better. As can be seen from the figure, both CG and UG can generate solutions with lower objective values.

\begin{itemize}
\item{\textbf{CG with conditional DM} can generate offspring solutions closer to the PF than SBX and UG by introducing performance gradient information in the denoising process, i.e., it \textbf{has stronger convergence performance}.}
\item{\textbf{UG with unconditional DM} does not have as strong convergence performance as CG, but since it does not rely on the gradient of the aggregated objective values to guide the denoising process, the solutions generated by UG will \textbf{have stronger diversity}.}
\end{itemize}

\subsection{The Advantages of Entropy Weighting Method}

The entropy weighting method, serving as an adaptive weight allocation approach, assigns weights based on the information entropy of objectives. This method reduces subjectivity in weighting gradients for different objective values, and \textbf{ensures that the algorithm places more emphasis on objectives with rich information content}, thereby achieving \textbf{better performance in multi-objective optimization problems}.

\begin{figure*}[h]
\vskip 0.2in
\begin{center}
% Adjust the widths as necessary to fit side by side
\includegraphics[width=0.50\textwidth]{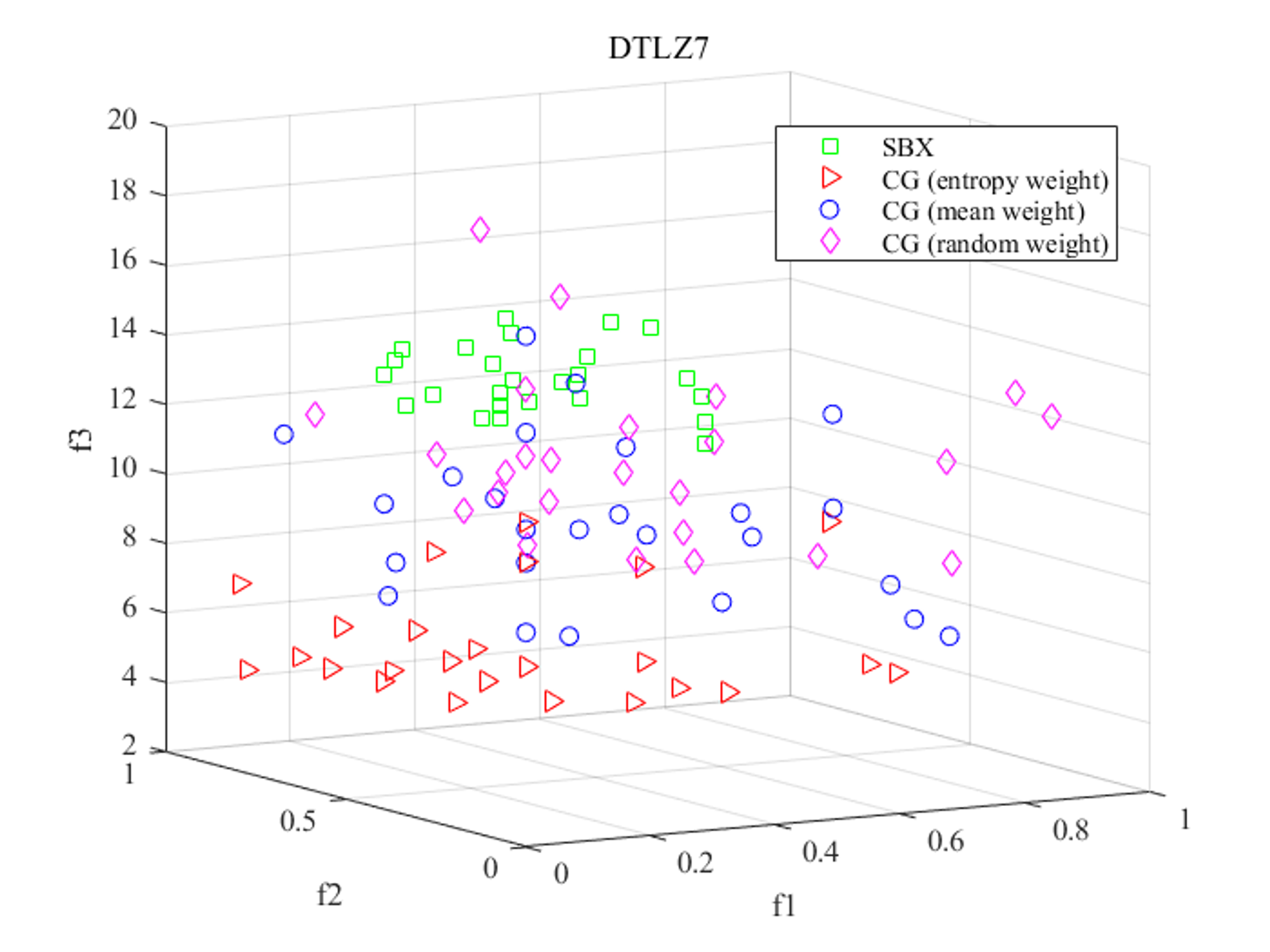}
\caption{The objective values of newly generated solutions obtained by GA's SBX method, CG(entropy weight), CG(mean weight) and CG(random weight). ($d = 20$, $FE = 25$)}
\label{ew_mw_rw_dtlz7}
\end{center}
\vskip -0.2in
\end{figure*}

To show the advantages of entropy weighting method, we visualize the offspring to compare the convergence performance of SBX and CG under three different weighting methods: entropy weighting, mean weighting, and random weighting. Figure \ref{ew_mw_rw_dtlz7} illustrates the objective values of all newly generated solutions at FE = 25. As can be seen from the figure, CG with weights can generate solutions with lower objective values across multiple objectives. Moreover, \textbf{CG with entropy weighted gradient has better convergence performance} compared to others. This is because CG (entropy weight) pay close attention to objectives that are relatively more important (with higher information entropy) during the generation of offspring. The offspring produced by it exhibit \textbf{stronger convergence} compared to those generated with CG (mean weight) and CG (random weight).

\subsection{The Advantages of CDM Over a Gradient-Based Method With Multiple Restarts}

In this section, we demonstrate through experiments the advantages of our proposed CDM-PSL compared to the gradient-based method (GM) with multiple restarts. The main advantage of CDM over gradient-based methods is that CDM is based on \textbf{population-wide information} to generate new offsprings, while some other general optimization methods are based on \textbf{individual solutions} to generate new offsprings and they \textbf{ignore the connections between solutions}, which makes CDM more advantageous in maintaining population diversity \cite{lin2022pareto}.

Moreover, gradient-based methods are more likely to \textbf{get stuck in local optima or saddle points} and fail to explore the global optimum, and \textbf{they are sensitive to the choice of the initial point and the step size} \cite{jin2021nonconvex}. While the problem of them falling into a local optimum can be mitigated to some extent by restart, this can lead to significant time overhead when dealing with high dimensional problem. However, \textbf{DM can model complex distributions of solutions}, and generate more diverse solutions by introducing random noise.

To compare the performance of CDM and GM through experiments, we implemented a GM with multiple restarts, with the following specific settings:

\begin{itemize}
    \item The objective value gradient of the current solution is obtained in the same manner as in CG.
    \item The weighting of the gradient adopts the entropy weighting method, which is the most effective in CG.
    \item SDE is used to select good solutions for the next generation of offspring through gradient methods.
    \item Data augmentation is performed on the selected solutions (starting points for the gradient method), including adding Gaussian noise/random perturbations, etc.
\end{itemize}

\begin{figure*}[h]
\vskip 0.2in
\begin{center}
% Adjust the widths as necessary to fit side by side
\includegraphics[width=0.50\textwidth]{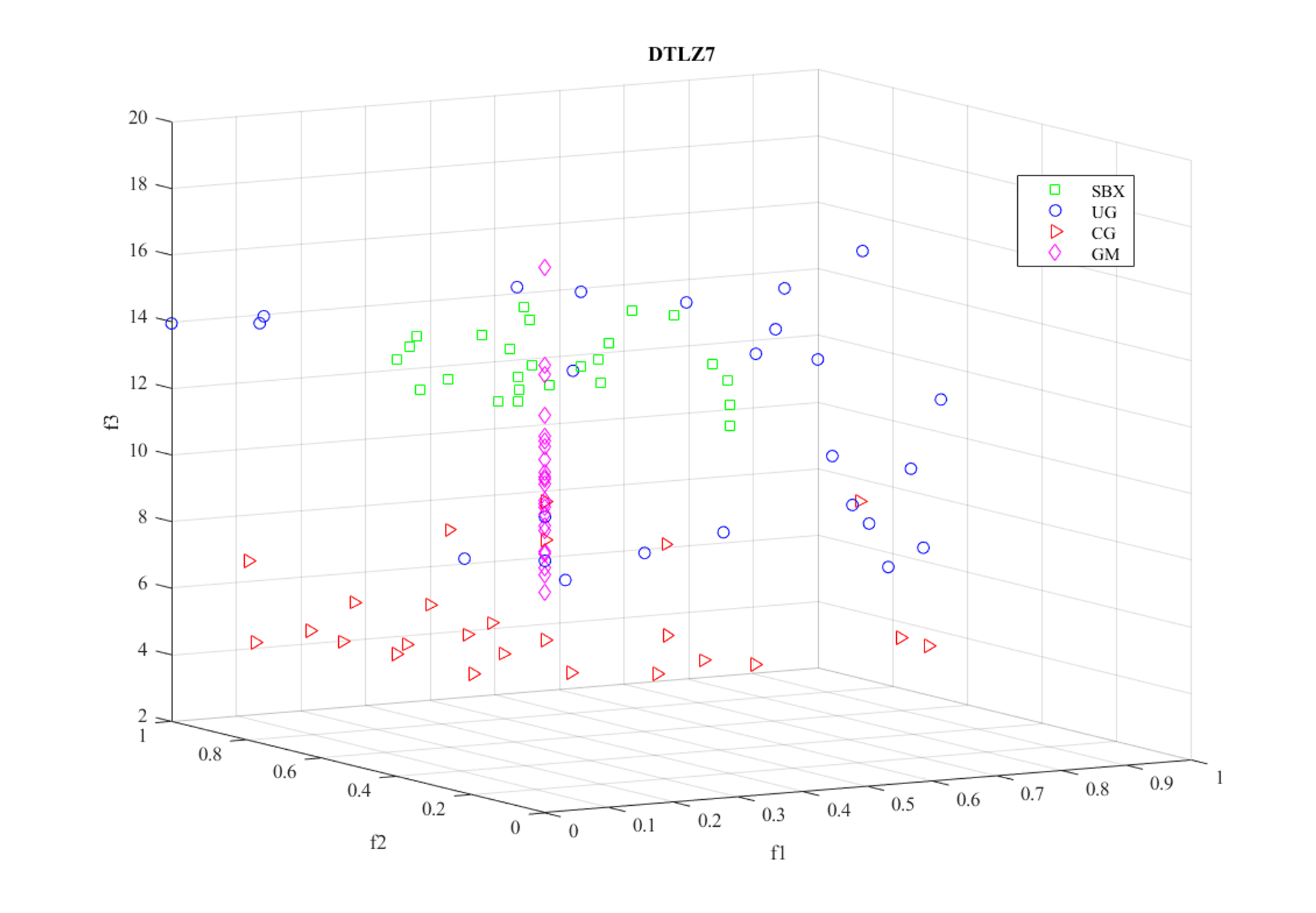}
\includegraphics[width=0.48\textwidth]{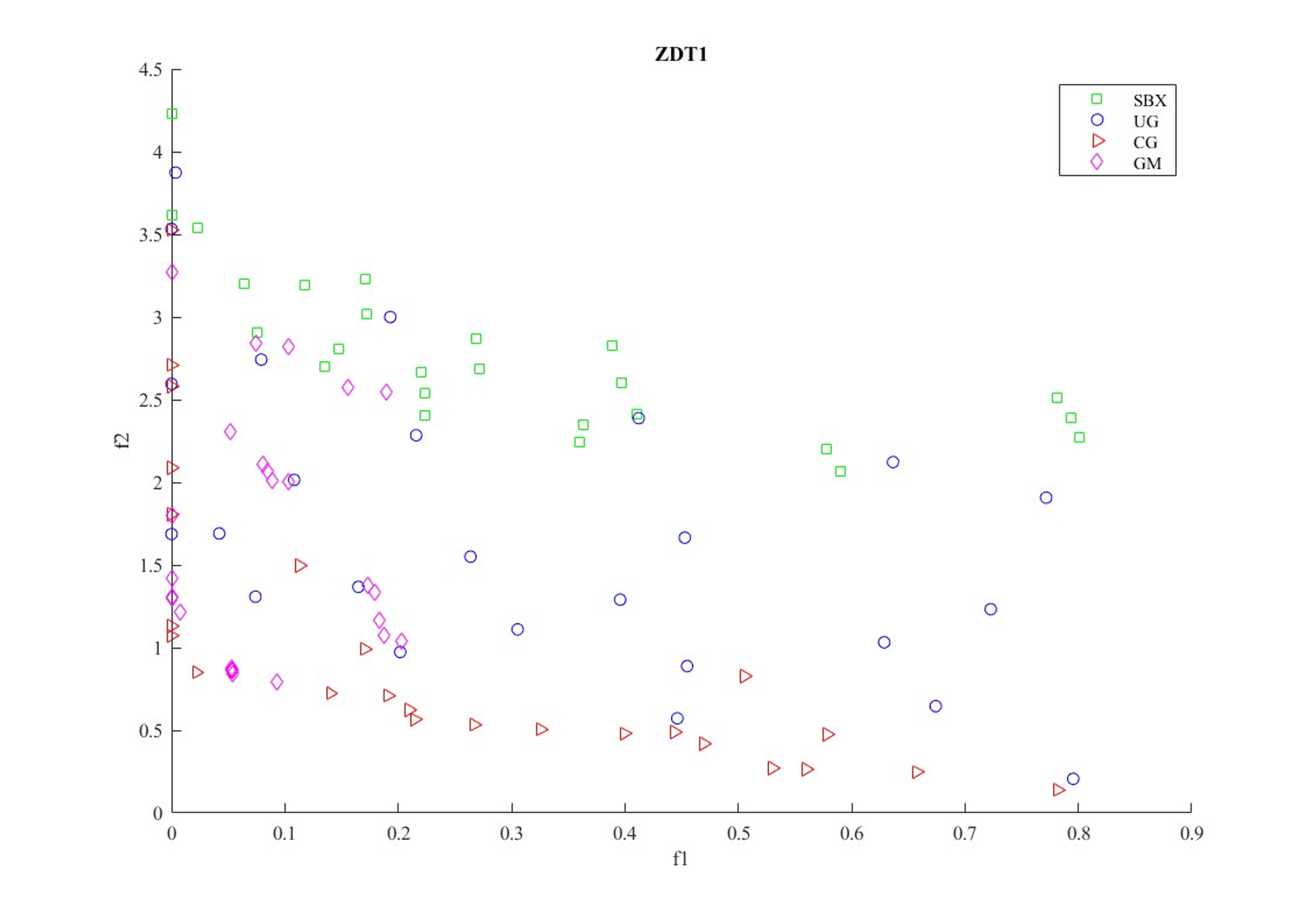}
\caption{The objective values of newly generated solutions obtained by GA's SBX method, UG, CG and GM. ($d = 20$, $FE = 25$)}
\label{gm-dtlz7-zdt1}
\end{center}
\vskip -0.2in
\end{figure*}

The experimental results are shown in Figure \ref{gm-dtlz7-zdt1} ($d=20, FE=25$), which validate the advantages of CDM over GM with multiple restarts by visualizing the offspring. GM generates solutions that are more likely to fall into local optimality, while CDM can maintain better diversity.

\begin{figure*}[h]
\vskip 0.2in
\begin{center}
\centerline{\includegraphics[width=0.65\textwidth]{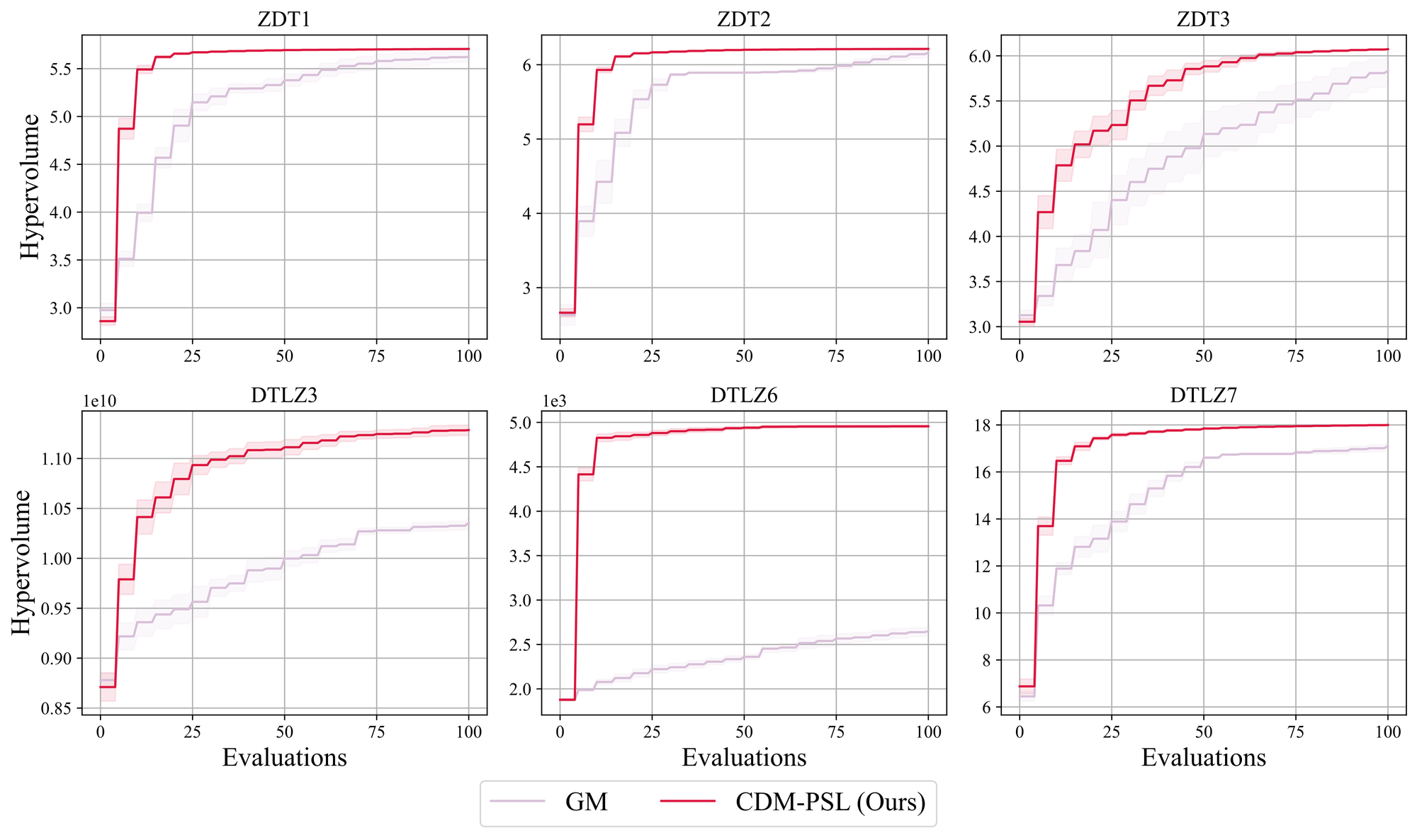}}
\caption{The HV results of CDM-PSL and GM ($d=20$).}
\label{gm-hv}
\end{center}
\vskip -0.2in
\end{figure*}

Figure \ref{gm-hv} demonstrates that \textbf{CDM, compared with GM, can generate solutions with better convergence and diversity} by comparing the HV values of both.

\subsection{Results on Synthetic Benchmark Problems with Various Decision Space Scales}

\begin{figure*}[ht]
\vskip 0.2in
\begin{center}
\centerline{\includegraphics[width=0.65\textwidth]{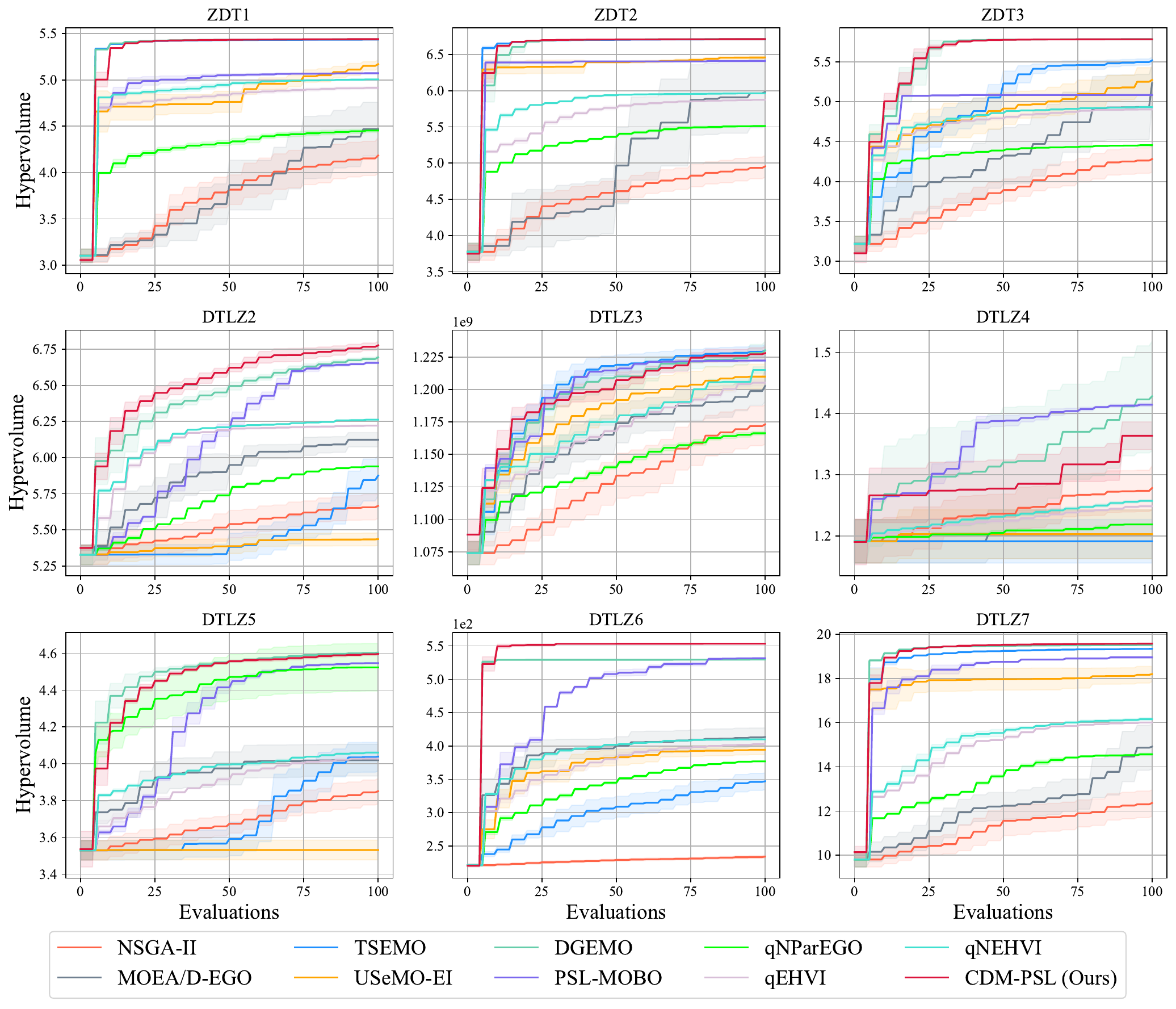}}
\caption{The HV results of 10 algorithms, evaluated on synthetic test functions and real-world problems ($d=10$).}
\label{final_hv-d10}
\end{center}
\vskip -0.2in
\end{figure*}

\begin{figure*}[ht]
\vskip 0.2in
\begin{center}
\centerline{\includegraphics[width=0.65\textwidth]{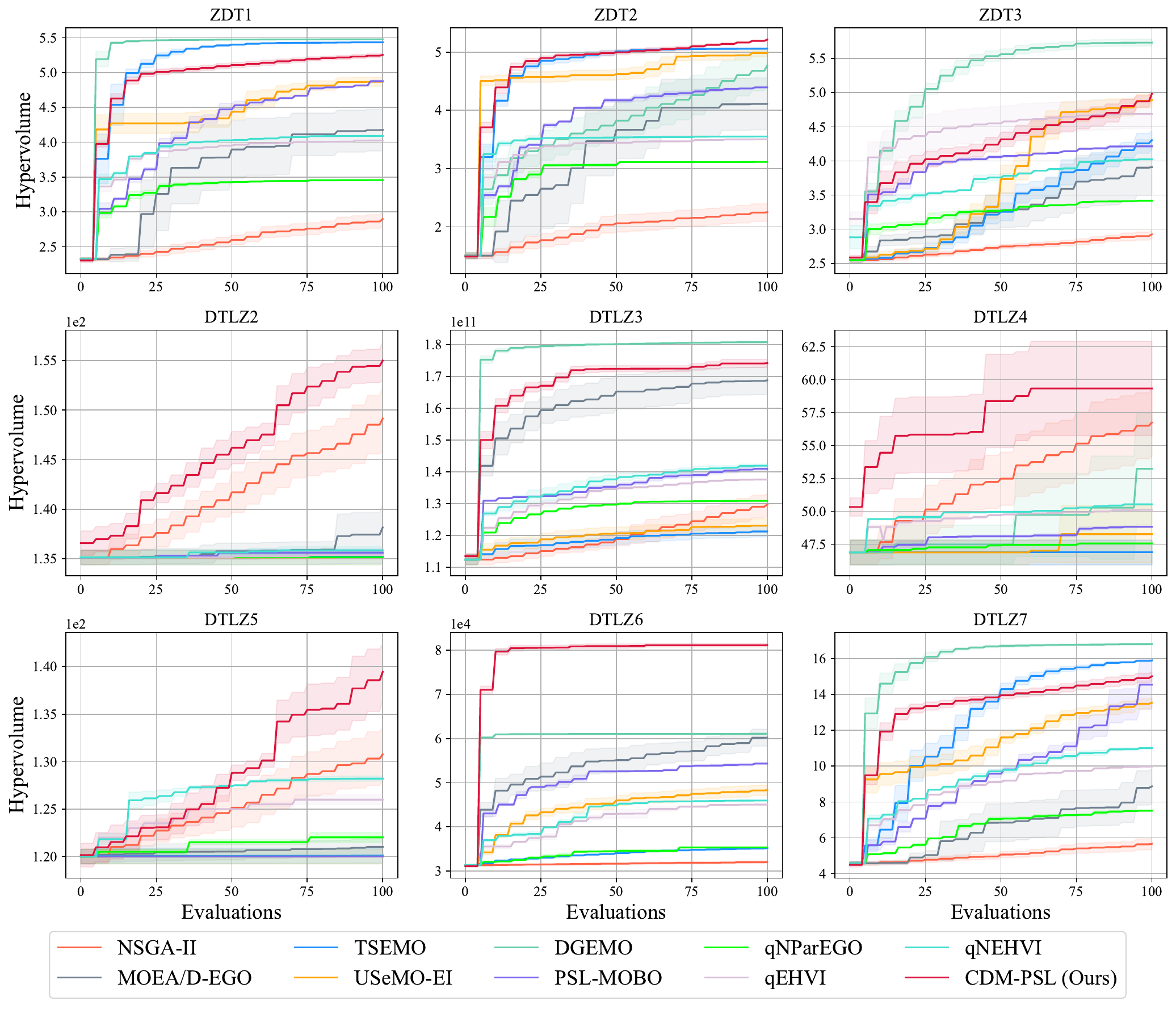}}
\caption{The HV results of 10 algorithms, evaluated on synthetic test functions and real-world problems ($d=50$).}
\label{final_hv-d50}
\end{center}
\vskip -0.2in
\end{figure*}

In this section, we broaden our experimental scope to encompass problems within 10-dimensional decision spaces, maintaining the same experimental parameters as before, including a batch size of 5, 20 iterations, and an initial sample size of 100. The outcomes, illustrated in Figure \ref{final_hv-d10} and \ref{final_hv-d50}, also demonstrate that, overall, CDM-PSL is superior to the methods compared.

Specifically, when the dimensionality of decision variables is set to 10, CDM-PSL exhibits highly competitive performance in almost all problems compared to the benchmark methods. Furthermore, when the dimensionality of decision variables reaches 50, our proposed method achieves notably superior performance on DTLZ2, DTLZ4, DTLZ5, and DTLZ6. However, the performance of CDM-PSL shows some degradation on ZDT3, DTLZ3, and DTLZ7. As indicated by the characteristics of benchmark problems in Table 2, these problems all possess a multimodal landscape feature. This phenomenon occurs because our proposed Pareto set learning approach generates new offspring based on the distribution of the current population. In these three types of problems, the Pareto optimal set is distributed across multiple regions, and with the increase in decision space dimensionality, the Pareto optimal sets of these multimodal landscape problems become more complex. As a result, our method might struggle with adequate exploration and exploitation on these types of problems at higher dimensions. In contrast, CDM-PSL excels in capturing unimodal Pareto sets (such as DTLZ2, DTLZ4, DTLZ5, and DTLZ6), particularly in high-dimensional decision spaces, showing a clear advantage over the methods compared.

\subsection{Results on Synthetic Benchmark Problems with Various Initial population size}

\begin{figure*}[h]
\vskip 0.2in
\begin{center}
\centerline{\includegraphics[width=0.65\textwidth]{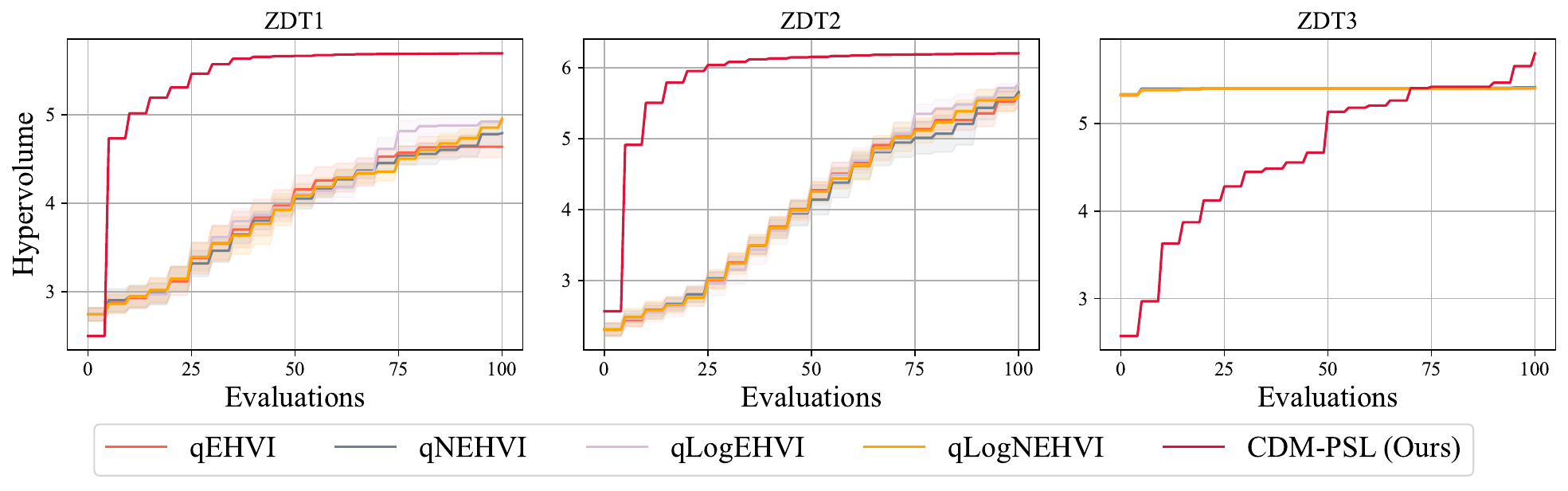}}
\caption{The HV results of 5 algorithms, with the initial population size of $2(d+1)$, where $d=20$.}
\label{diff-init}
\end{center}
\vskip -0.2in
\end{figure*}

In this section, we experiment with the performance of CDM-PSL under varying initial population sizes. We selected qEHVI \cite{daulton2020differentiable}, qNEHVI \cite{daulton2021parallel}, qLogEHVI \cite{ament2024unexpected}, and qLogNEHVI \cite{ament2024unexpected} as comparison algorithms. The initial population size for these algorithms were set according to their default settings, specifically $2(d+1)$, where $d$ represents the dimension of the solution. Figure \ref{diff-init} presents the results of these five algorithms on different problems. From the figure, it is evident that even with a smaller initial population size, CDM-PSL maintains good convergence performance in the early stages of the algorithm. Furthermore, compared to the other algorithms, CDM-PSL achieves higher HV values in the final evaluation.

\subsection{Comparation of Log HV Differences}

\begin{figure*}[h]
\vskip 0.2in
\begin{center}
\centerline{\includegraphics[width=0.86\textwidth]{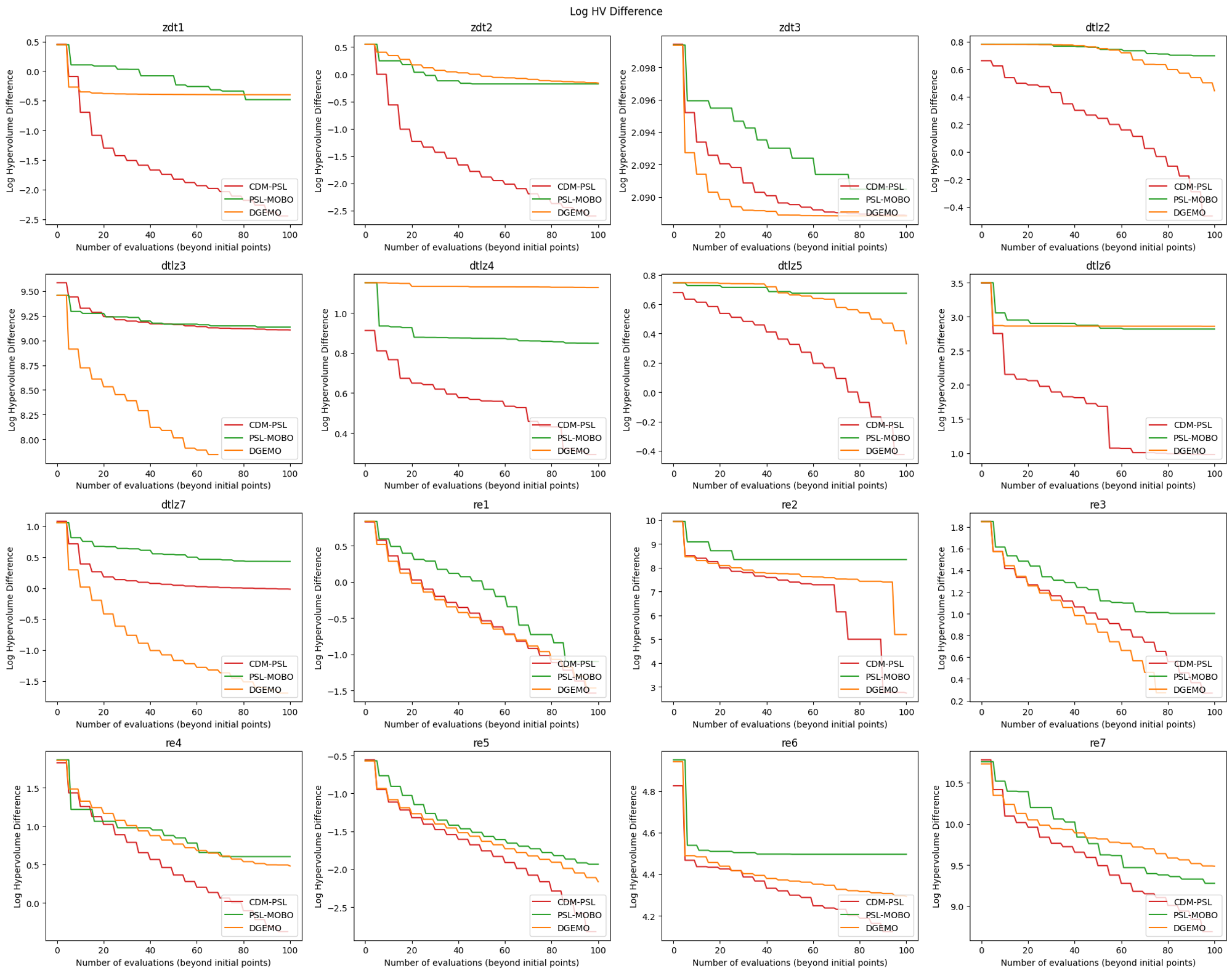}}
\caption{The log HV difference of 3 competitive algorithms, evaluated on synthetic test functions and real-world problems ($d=20$).}
\label{log-hv-diff}
\end{center}
\vskip -0.4in
\end{figure*}

In this section, we include the plot of log HV difference for all the test problems ($d=20$). Figure \ref{log-hv-diff} shows the plots of log HV difference for CDM-PSL (our method) alongside PSL-MOBO and DGEMO, which are two of the most competitive algorithms examined in our experiments. This comparison effectively highlights what fraction of the problem was actually solved and showcase the superior performance of CDM-PSL.

\subsection{Ablation Study with Limited FEs}

To analyze the convergence performance of our proposed method, we compare CDM-PSL with three variants: CDM-PSL without entropy weight (CDM-PSL w/o Weight), CDM-PSL without conditional generation (CDM-PSL w/o Condition), CDM-PSL without diffusion model (CDM-PSL w/o DM). As can be seen from the Table \ref{convergence-hv}, the convergence performance of CDM-PSL is impaired when any of the components is ablated. Notably, incorporating the diffusion model to generate offspring significantly improves the model's performance with a minimal number of function evaluations (specifically, 25), enabling it to achieve higher HV values within a limited set of evaluations.

\begin{table}[!htbp]  
  \centering
  \caption{Comparison of HV Values for CDM-PSL and its variants at 25 FEs.}
  \label{convergence-hv}
  \setlength{\tabcolsep}{2pt}
  \setlength{\abovecaptionskip}{0cm}
  \scriptsize
  \scalebox{1.0}{
    \begin{tabular}{l|cc|cc|cc|cc|cc|cc}
      \toprule
      & \multicolumn{2}{c|}{ZDT1} & \multicolumn{2}{c|}{ZDT2} & \multicolumn{2}{c|}{ZDT3}
      & \multicolumn{2}{c|}{DTLZ3} & \multicolumn{2}{c|}{DTLZ6} & \multicolumn{2}{c}{DTLZ7}\\
      Method & HV $\uparrow$ & Gap $\downarrow$ & HV $\uparrow$ & Gap $\downarrow$ & HV $\uparrow$ & Gap $\downarrow$ 
             & HV $\uparrow$ & Gap $\downarrow$ & HV $\uparrow$ & Gap $\downarrow$ & HV $\uparrow$ & Gap $\downarrow$ \\
      \midrule
      CDM-PSL w/o Weight     & 4.9542e+00 & 12.42\% & 5.4062e+00 & 12.14\% & 4.2659e+00 & 17.49\% 
                             & 1.0391e+10 & 3.74\% & 4.3509e+03 & 10.46\% & 1.4604e+01 & 16.22\% \\
      CDM-PSL w/o Condition  & 4.9587e+00 & 12.34\% & 5.1008e+00 & 17.10\% & 4.1583e+00 & 19.57\% 
                             & 1.0420e+10 & 3.47\% & 4.2265e+03 & 13.02\% & 1.4622e+01 & 16.12\% \\
      CDM-PSL w/o DM         & \underline{3.6577e+00} & \underline{34.35\%} & \underline{3.7327e+00} &
                             \underline{39.34\%} & \underline{3.7185e+00} & \underline{28.08\%}
                             & \underline{9.6465e+09} & \underline{10.64\%} & \underline{2.2088e+03} & \underline{54.54\%} & \underline{9.4968e+00} & \underline{45.52\%} \\
      \midrule
      CDM-PSL                & \textbf{5.6570e+00} & \textbf{0.00\%} & \textbf{6.1533e+00} & \textbf{0.00\%} & \textbf{5.1703e+00} & \textbf{0.00\%} & \textbf{1.0795e+10} & \textbf{0.00\%} & \textbf{4.8592e+03} & \textbf{0.00\%} & \textbf{1.7432e+01} & \textbf{0.00\%} \\
      \midrule
    \end{tabular}
  }
\end{table}

\subsection{Pareto Fronts Obtained by CDM-PSL on Real-World Problems}

\begin{figure*}[h]
\vskip 0.2in
\begin{center}
% \centerline{\includegraphics[width=0.76\textwidth]{fig/ablation.pdf}}
    \subfigure{\begin{minipage}[h]{0.25\textwidth}
        \includegraphics[width=1\textwidth]{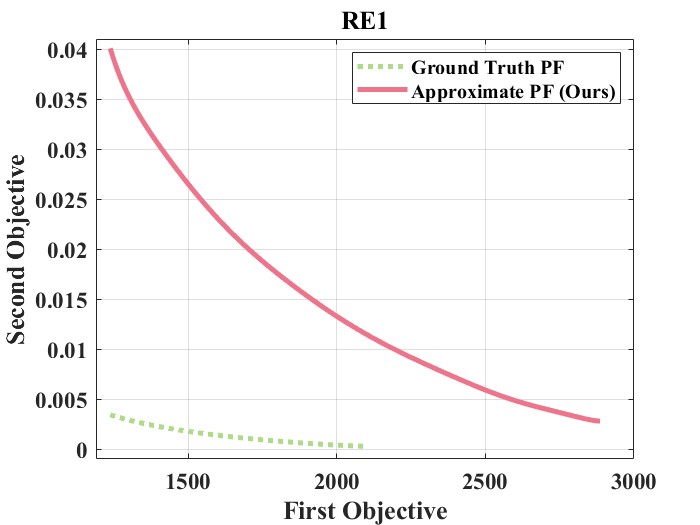}
    \end{minipage}}
    \subfigure{\begin{minipage}[h]{0.25\textwidth}
        \includegraphics[width=1\textwidth]{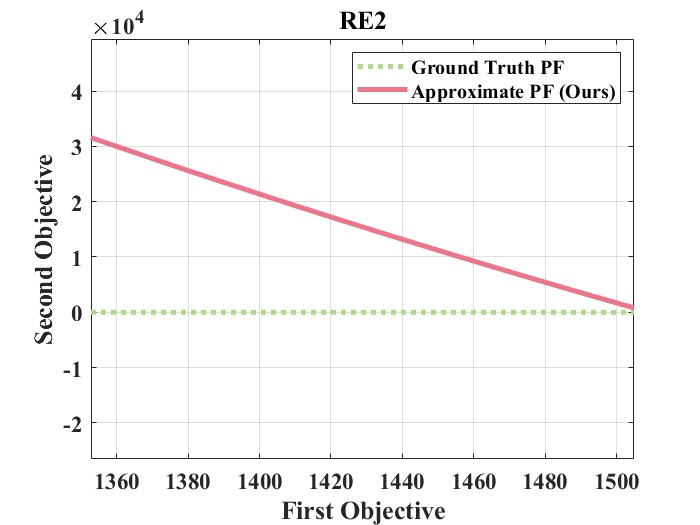}
    \end{minipage}}
    \subfigure{\begin{minipage}[h]{0.25\textwidth}
        \includegraphics[width=1\textwidth]{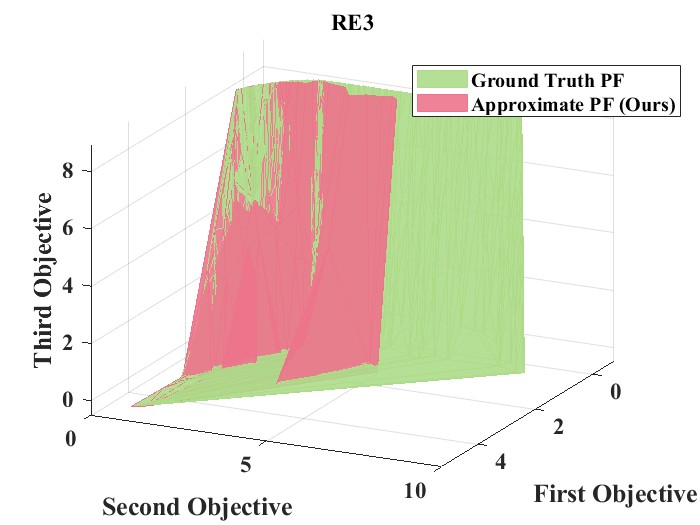}
    \end{minipage}}
    \subfigure{\begin{minipage}[h]{0.25\textwidth}
        \includegraphics[width=1\textwidth]{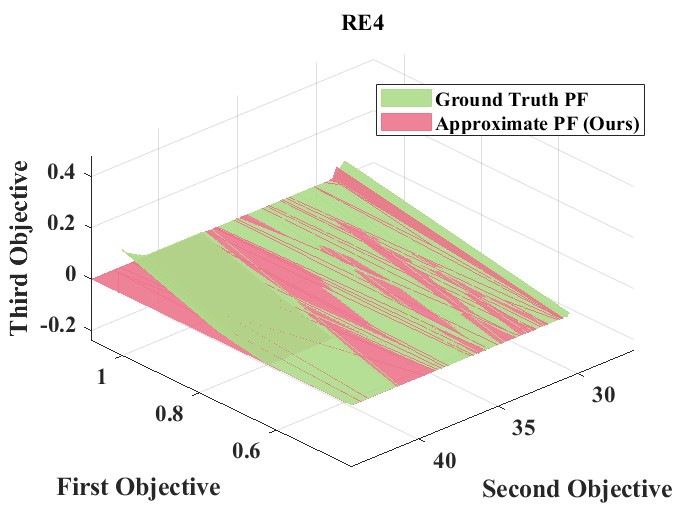}
    \end{minipage}}
    \subfigure{\begin{minipage}[h]{0.25\textwidth}
        \includegraphics[width=1\textwidth]{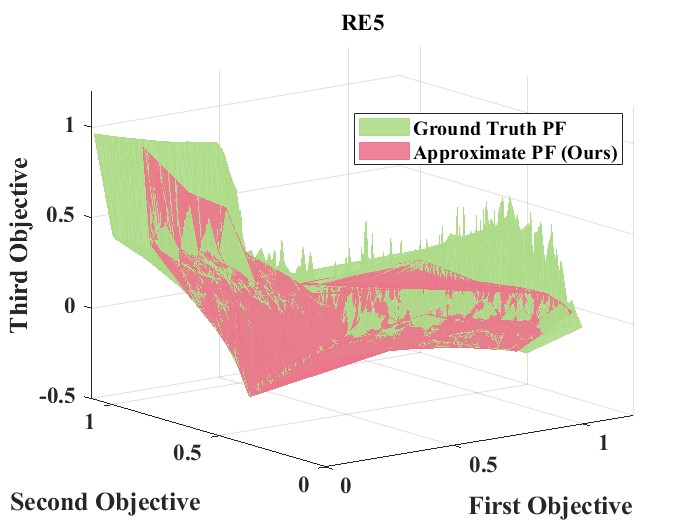}
    \end{minipage}}
    \subfigure{\begin{minipage}[h]{0.25\textwidth}
        \includegraphics[width=1\textwidth]{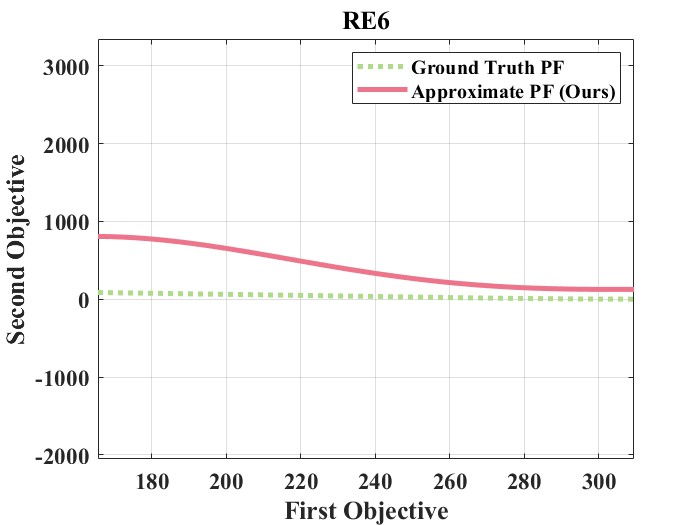}
    \end{minipage}}
    \subfigure{\begin{minipage}[h]{0.25\textwidth}
        \includegraphics[width=1\textwidth]{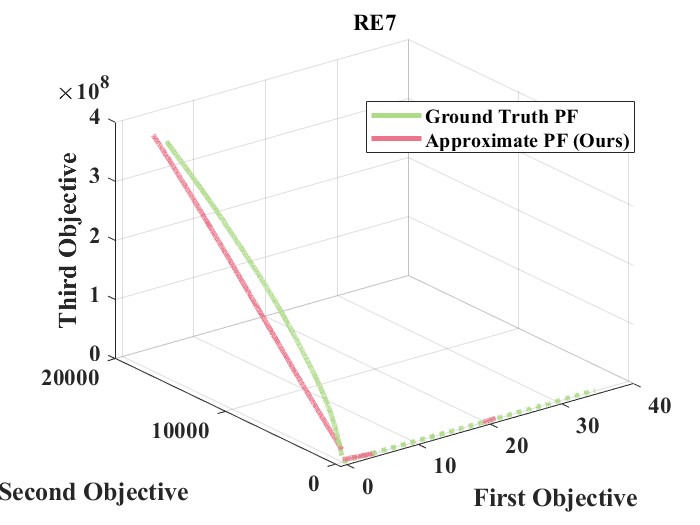}
    \end{minipage}}
\caption{Approximate Pareto fronts obtained by CDM-PSL in real-word problems. ($d = 20$, $FE = 100$)}
\label{pf-rebuttal}
\end{center}
\vskip 0.2in
\end{figure*}

In this section, we add more figures of the final solutions obtained by CDM-PSL in Real-World Problems. As can be seen from the figures \ref{pf-rebuttal}, the proposed CDM-PSL can effectively modeling the distribution of the PS, despite that the Pareto fronts of the real-world test problems RE1-RE7 are complicated.

\subsection{Parameter Analysis}
\subsubsection{Performance with Different Learning Rates}
\begin{figure*}[h]
\vskip 0.2in
\begin{center}
\centerline{\includegraphics[width=0.65\textwidth]{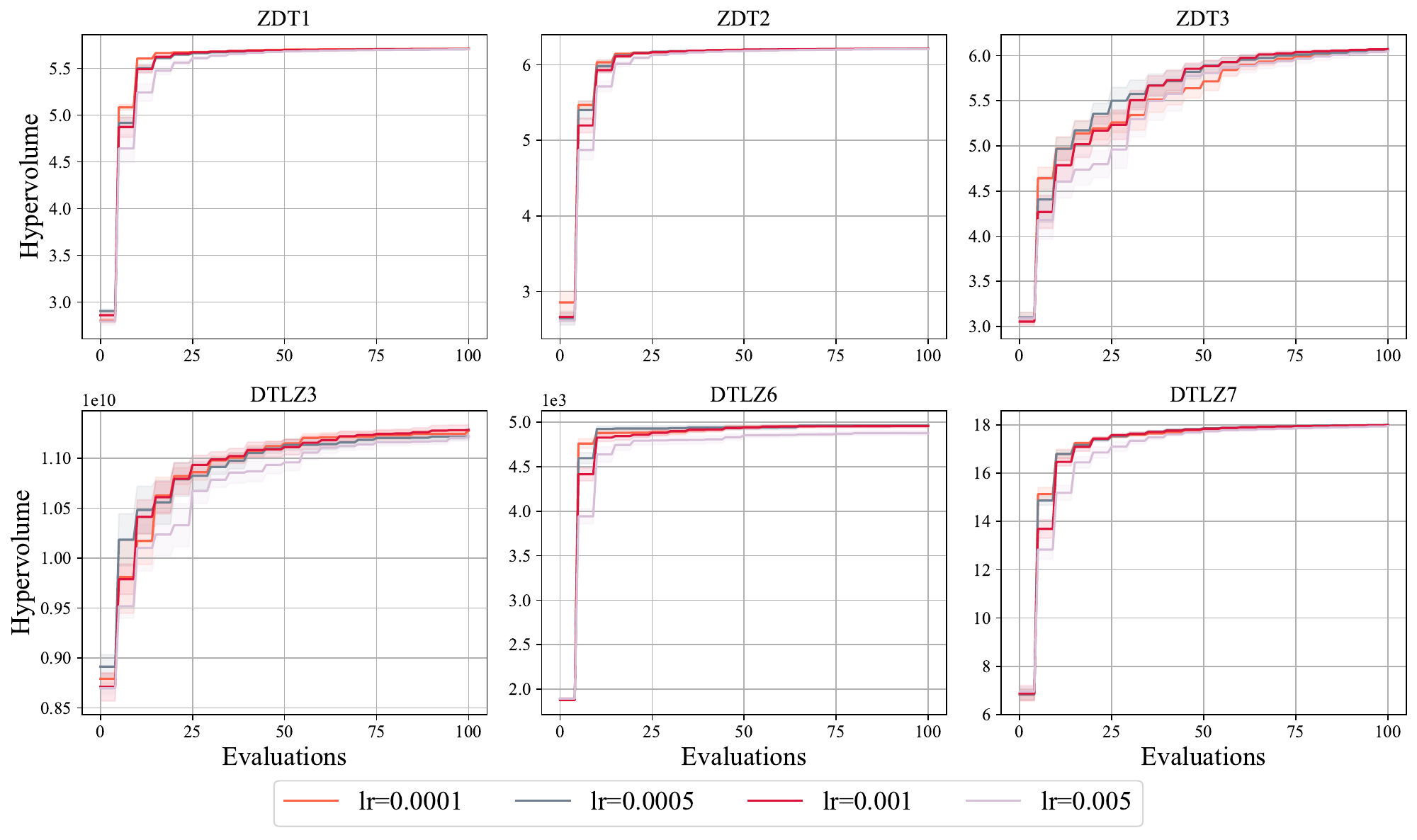}}
\caption{The HV results of CDM-PSL at different learning rate ($d=20$).}
\label{parameter-lr}
\end{center}
\vskip -0.2in
\end{figure*}

In this section we compare the performance of CDM-PSL under different learning rate settings to validate the sensitivity of our proposed method to this parameter. Figure \ref{parameter-lr} illustrates the impact of varying learning rates on the performance of CDM-PSL, notably with $lr=0.001$ being the default setting of our proposed method. The results depicted in Figure \ref{parameter-lr} indicate that there are no significant differences in the performance of CDM-PSL across different learning rates.

\subsubsection{Study of CG and UG ratios in CDM-PSL}
\begin{figure*}[h]
\vskip 0.2in
\begin{center}
\centerline{\includegraphics[width=0.65\textwidth]{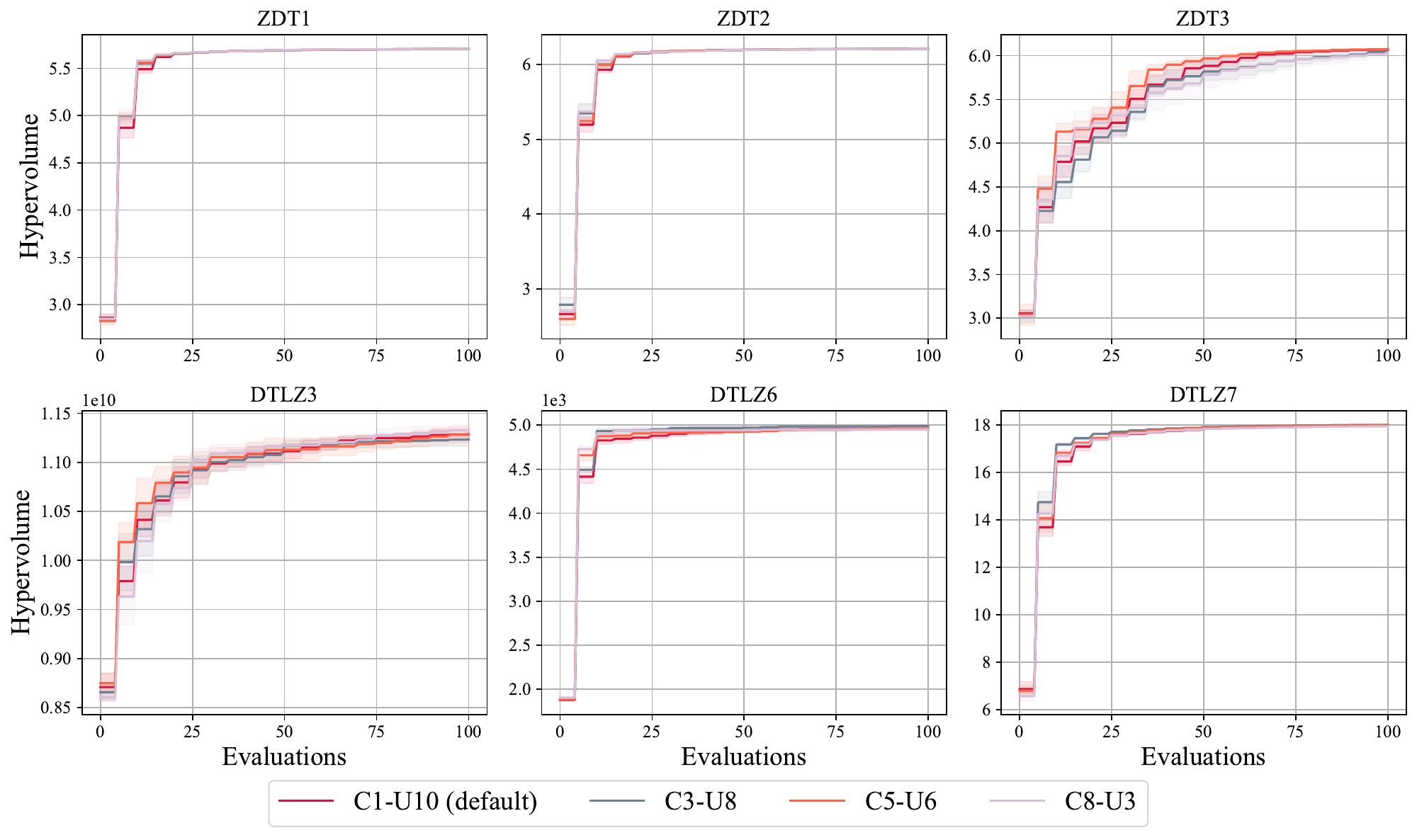}}
\caption{The HV results of CDM-PSL with different CG and UG ratios ($d=20$).}
\label{parameter-ugcg}
\end{center}
\vskip -0.2in
\end{figure*}

In this section we exploring the impact of varying ratios between conditional generation and unconditional generation in CDM-PSL. Figure \ref{parameter-ugcg} displays the HV values for CDM-PSL with different CG and UG ratios across multiple problems. Here, "C1-U10" signifies a CG to UG ratio of 1 to 10, which is the default setting for CDM-PSL; "C3-U8" indicates a ratio of 3 to 8; and so forth. From the figure, it is observable that incrementing the proportion of CG from the default configuration marginally improves the algorithm's early convergence performance, as evidenced by a slightly elevated HV value up to the 25th function evaluation. However, with a sufficient number of function evaluations, the DM has been able to learn the distribution of the good solutions more accurately. Consequently, the optimal final HV value is achievable with a minimal CG ratio and a predominately higher experimental group UG ratio. Therefore, we opt for a CG to UG ratio of 1 to 10, which significantly reduces the algorithm's time overhead without compromising its final performance.

\subsection{Time overhead of CDM-PSL}

We choose ZDT1 and DTLZ2 to discuss the time overhead of CDM-PSL, the number of decision variables is 10 and 20 respectively, the maximum number of function evaluations is set to 120, and the parameters for training CDM are consistent with the default settings mentioned in the manuscript.

Table \ref{time-overhead-comp} lists the time overhead of each part of the CDM-PSL, including training, CG, UG and others. We choose ZDT1 and DTLZ2 to discuss the time overhead of the algorithms, the number of decision variables is 10 and 20 respectively, the maximum number of function evaluations is set to 120, and the parameters for training CDM are consistent with the default settings mentioned in the manuscript.

As we can see from the table, the time overhead of training DM is a relatively modest fraction of the total time overhead of CDM-PSL. CG takes longer to generate new candidate solutions compared to UG. This significant difference is attributed to CG's reliance on a surrogate model to calculate the objective values during the denoising process, which is essential for obtaining the gradient information needed to guide the denoising. However, the additional time overhead imposed by CDM is acceptable compared to the rest of the algorithm.

In the far right column of the table, we present the time overhead associated with one of the latest SOTA algorithms, PSL-MOBO. A comparison between the total time overhead of our method and that of PSL-MOBO indicates that the additional time overhead incurred by CDM is within acceptable limits.

\begin{table}[!htbp]  
  \centering
  \caption{Time overhead of algorithm components in CDM-PSL.}
  \label{time-overhead-comp}
  \setlength{\tabcolsep}{2pt}
  \setlength{\abovecaptionskip}{0cm}
  \scriptsize
  \scalebox{1.0}{
    \begin{tabular}{l|cc|cc|cc|cc|c|c}
      \toprule
      & \multicolumn{2}{c|}{Training} 
      & \multicolumn{2}{c|}{CG} 
      & \multicolumn{2}{c|}{UG}
      & \multicolumn{2}{c|}{Other} 
      & {CDM-PSL}
      & {PSL-MOBO}\\
      Instance 
      & Time(s) & Proportion 
      & Time(s) & Proportion 
      & Time(s) & Proportion 
      & Time(s) & Proportion 
      & Time(s)
      & Time(s)\\
      \midrule
      ZDT1 ($d=10$)     
      & 12.5578 & 21.13\% 
      & 9.5404 & 16.05\% 
      & 1.7426 & 2.93\% 
      & 35.5903 & 59.88\% 
      & 59.4311
      & 32.5015\\
      ZDT1 ($d=20$)  
      & 12.5281 & 22.11\% 
      & 12.5824 & 22.20\% 
      & 1.7389 & 3.07\% 
      & 29.8183 & 52.62\% 
      & 56.6677
      & 38.7551\\
      DTLZ2 ($d=10$)  
      & 12.5111 & 14.32\% 
      & 12.9713 & 14.85\% 
      & 1.7146 & 1.96\% 
      & 60.1494 & 68.86\% 
      & 87.3464
      & 55.2474\\
      DTLZ2 ($d=20$)  
      & 12.5622 & 13.33\% 
      & 18.2628 & 19.38\% 
      & 1.7364 & 1.84\% 
      & 61.6794 & 65.45\% 
      & 94.2408
      & 58.8312\\
      \midrule
    \end{tabular}
  }
\end{table}

\section{More Explanations About the Motivation of the Proposed Method}

\subsection{Motivation of using DM in PSL-based MOBO}
When addressing EMOPs, the performance of PSL methods might deteriorate significantly due to the \textbf{complexity of distributions} of Pareto optimal solutions and the \textbf{limitation of function evaluations}. This poses a significant challenge to existing PSL algorithms \cite{lin2022pareto}. Unlike existing methods that directly learn the distribution of the PS in a single-step way, \textbf{complex distributions of solutions can be learned step by step with the aid of DM}. The success of DM in various domains (e.g. computer vision) shows promise for its application in PSL in EMOPs.

\subsection{Detailed motivation}
Using DM for MOBO is not trival, because one should take into account the following two aspect: how to speed up the searching process and how to balance convergence and diversity. Having this in mind, we proposed a PSL method with composite diffusion models. Specifically, the proposed CDM-PSL has the following features:

\begin{itemize}
    \item CG with conditional DM is aimed at \textbf{better convergence} since it introduces gradient information to guide the searching process.
    \item UG with unconditional DM is designed for generating \textbf{diverse} solutions with low cpu-time cost.
    \item The entropy weighting method, serving as an objective aggregation approach for offering performance gradient information, is introduced to assign weights based on the information entropy of objectives. This method reduces subjectivity in weighting gradients for different objective values, and \textbf{ensures that the algorithm places more emphasis on objectives with rich information content}. Figure \ref{ew_mw_rw_dtlz7} shows that CG with entropy weighting method can generate better solutions than CG with mean weighting or random weighting.
\end{itemize}

\section{Limitations of CDM-PSL}

In this section, we will discuss some limitations of CDM-PSL. One limitation of our approach is that CG will introduce additional time overhead. As shown in Table \ref{time-overhead-comp}, conditional generation increases the time overhead of generating candidate solutions to some extent, which leads to the limitations of the method in terms of excessive time overhead when applied to very high dimensional problems. Another limitation of our work is that our experiments were not conducted on real-world problems with very high dimensionality. While we have conducted experiments on seven real-world engine design problems, RE1-RE7, which can provide some evidence of the superiority of CDM-PSL in facing real-world problems, these test cases do not represent the complexity and scale of extremely high-dimensional industrial problems. We will explore the algorithm's performance on super large problems like real industrial design in future work.

\end{document}